\documentclass[letterpaper]{article} 
\usepackage{aaai25}  
\usepackage{times}  
\usepackage{helvet}  
\usepackage{courier}  
\usepackage[hyphens]{url}  
\usepackage{graphicx} 
\usepackage{natbib}  
\usepackage{caption} 
\usepackage{algorithm}

\usepackage{multirow}
\usepackage{caption}
\usepackage{subcaption}
\usepackage{makecell}
\usepackage{listings}
\usepackage{xspace}

\usepackage{algorithm}
\PassOptionsToPackage{noend}{algpseudocode}
\usepackage{algpseudocode}
\errorcontextlines\maxdimen

\newcommand{\sol}{{\em UnifiedNN}\xspace}
\newcommand{\eg}{{\em e.g.,}\ }

\usepackage{tikz}
\usepackage{amsmath}

\usepackage{newfloat}
\usepackage{listings}
\DeclareCaptionStyle{ruled}{labelfont=normalfont,labelsep=colon,strut=off} 
\floatstyle{ruled}
\newfloat{listing}{tb}{lst}{}
\floatname{listing}{Listing}
\title{\sol: Efficient Neural Network Training on the Cloud}
\author{
    Sifat Ut Taki\textsuperscript{\rm 1},
    Arthi Padmanabhan\textsuperscript{\rm 2},
    Spyridon Mastorakis\textsuperscript{\rm 1}
}
\affiliations{
    \textsuperscript{\rm 1}University of Notre Dame\\
    \textsuperscript{\rm 2}Harvey Mudd College\\

    staki@nd.edu, arpadmanabhan@g.hmc.edu, mastorakis@nd.edu
}

\usepackage{bibentry}

\begin{document}

\maketitle

\begin{abstract}
Nowadays, cloud-based services are widely favored over the traditional approach of locally training a Neural Network (NN) model. Oftentimes, a cloud service processes multiple requests from users--thus training multiple NN models concurrently. However, training NN models concurrently is a challenging process, which typically requires significant amounts of available computing resources and takes a long time to complete. In this paper, we present \sol to effectively train multiple NN models concurrently on the cloud. \sol effectively ``combines'' multiple NN models and features several memory and time conservation mechanisms to train multiple NN models simultaneously without impacting the accuracy of the training process. 
Specifically, \sol merges multiple NN models and creates a large singular unified model in order to efficiently train all models at once. We have implemented a prototype of \sol in PyTorch and we have compared its performance with relevant state-of-the-art frameworks. Our experimental results demonstrate that \sol can reduce memory consumption by up to 53\% and training time by up to 81\% when compared with vanilla PyTorch without impacting the model training and testing accuracy. Finally, our results indicate that \sol can reduce memory consumption by up to 52\% and training time by up to 41\% when compared to state-of-the-art frameworks when training multiple models concurrently.

\end{abstract}

\section{Introduction}
Over the last few years, there has been notable momentum in the adoption of Deep Learning (DL) applications. As a result, training Neural Networks (NN) efficiently has been a major challenge, since the training process is computationally expensive. This has increased the demand for access to powerful computing resources~\cite{nn_state, nn_state2}. To accelerate the training process of an NN model, modern accelerator hardware, such as Graphics Processing Units (GPUs) and Tensor Processing Units (TPUs), is used, which significantly reduces the overall training time by parallelizing computation~\cite{nn_hardware}. However, this hardware is expensive; as such, cloud-based solutions have emerged and users can resort to cloud infrastructures to train their models at a reasonable cost. Cloud-based Machine Learning (ML) solutions are more popular now than ever before. Major cloud service providers allow users to train and run NN models in their cloud infrastructures. Amazon SageMaker and Google Vertex AI are among the most popular cloud services that are available today.


As applications become more complex, NN models are increasing exponentially in size--requiring more computing resources for training. The model sizes have increased exponentially over the years, and it is forecasted that they will continue increasing in the future~\cite{nn_size}. As a consequence of the increasing model sizes, the training process is expected to become more and more resource-hungry. We argue that simply ``throwing more hardware to the problem'' may not be a sustainable way to move forward even for large-scale cloud providers, who have an abundance of financial and computing resources. To this end, frameworks that contribute to the effective utilization of the available computing resources during NN model training will be much needed~\cite{efficient_train1, efficient_train2, efficient_train3}.


In line with this argument, we introduce \sol, a framework to effectively train multiple NN models on the cloud by merging them together in this paper. 
\sol achieves effectiveness during training by merging multiple NN models, generating a hybrid model by combining the models from multiple users and training the hybrid model. \sol can achieve it without compromising the privacy of the users. As a result, multiple NN models are trained all together, reducing the overall usage of computing resources. In addition, \sol employs several systems-level mechanisms, such as loading deep learning libraries, transferring models to GPUs, and copying datasets \textbf{only once}, to reduce the overall memory usage and training time. 
\sol does not affect the training process or the accuracy of individual NN models, nor it leaks information from one model to another. Moreover, from the point of view of a user, the process is the same as it would have been if a model was trained separately on the cloud. Although \sol focuses on cloud computing environments, it can be deployed on any GPU server (local or on the cloud) that is used for NN model training. In this paper, our contribution is two-fold:


\begin{itemize}

    \item We present the design of \sol, which features mechanisms to combine multiple NN models into a hybrid model, train the hybrid model with different scheduling options, and separate the original models from the hybrid model once the training process is over. We also present an end-to-end operational workflow of \sol along with the memory and time-saving strategies that it employs.
    
    \item We present our prototype implementation of \sol using PyTorch and discuss the implementation of different components. We evaluate this prototype by comparing its performance against vanilla PyTorch and state-of-the-art frameworks. Our evaluation results demonstrate that \sol can reduce memory consumption by up to 53\% and training time by up to 81\% when compared with vanilla PyTorch without impacting the model training and testing accuracy. Our results also indicate that \sol can reduce memory consumption by up to 52\% and training time by up to 41\% when compared to state-of-the-art frameworks. 
    
    
\end{itemize}


\section{Related Work}
\label{sec:related}

Over the past few years, researchers have focused on optimizing the process of NN model training. As model architectures keep getting more extensive and more complex, this problem has become more important than ever. 

Zancato et. al. proposed a framework to predict the training time using the number of optimization steps required for a pre-trained model to converge to a certain loss value~\cite{zancato2020predicting}. However, this approach only works when fine-tuning a model and fails to predict the training time for models with random weight initialization. Predicting the GPU memory consumption when training an NN model with a given dataset is also a topic of interest. Some prediction models have been proposed over the years for the estimation of cost for an NN model. Canziani et. al. adopted a performance analyzer for deep learning~\cite{canziani2016analysis}. In their work, they analyzed the accuracy, memory footprint, parameters, operation count, inference time, and power consumption of an NN model. However, their approach was limited in terms of memory consumption analysis as they only considered a subset of the operations that contribute to the entire GPU memory consumption. 


Memory-efficient inference in edge-computing use cases is a major challenge. Model sharing is one of the common approaches, where outputs are shared across layers to save on memory usage~\cite{model_share1, model_share2}. PRETZEL~\cite{pretzel} reuses operators in traditional ML models, where parameters and layers are shared across multiple ML models. Multitask learning is another well-known approach where multiple models are trained simultaneously by sharing information with one another~\cite{multitask1, multitask2}. In video analytics use cases, frameworks, such as Chameleon~\cite{chameleon}, profile configurations and adopt the most effective configuration for maximum resource utilization.

To make the training phase efficient, several approaches have been recently proposed, which focus on memory efficiency. moDNN \cite{modnn} proposed by Chen et. al is another GPU memory optimization framework. moDNN adopts the idea of out-of-core algorithms and can automatically produce training code for a given memory budget. It optimizes memory footprint by offloading and prefetching the data. It can also support sub-batches to reduce batch size and memory allocation. ZCOMP~\cite{zcomp} is a cross-layer memory footprint reduction framework using vector extensions. ZCOMP is an instruction set architecture tailored for cross-layer communication. It optimizes DNN intermediate data streaming by fetching data as sequential or block sequential compression that eliminates random element retrieval. However, ZCOMP cannot access data elements randomly as both compression and expansion need to be performed
vector-by-vector sequentially for each piece of the dataset. Finally, Zico~\cite{zico} is another GPU memory optimization framework for concurrent DNN model training. Zico provides a memory management scheme that allows NN models to share GPU memory resources. Zico shares intermediate data generated during co-located training jobs to reduce the total memory footprint. However, as the set size of a job increases, the framework struggles with the memory-sharing scheme.

PipeDream~\cite{pipedream} schedules forward and backward passes on different mini-batches concurrently on different hardware workers. To train a large NN model across multiple accelerators efficiently, the authors have employed a pipelining and weight gradient coalescing strategy that uses double buffering of weight for low memory consumption and high throughput. They achieved $6.9\times$ acceleration when training a large language model. However, this framework has an overhead as it requires profiling before the model can be pipelined, which takes additional time. Moreover, there is a slight impact on the accuracy of the trained model. DropIT~\cite{chen2022dropit} introduces a technique where the intermediate cached tensors are dropped from the memory. The authors showed that the unnecessary cached tensor elements can be dropped without impacting the training of an NN model by up to 90\% for convolution and fully connected layers. They also proposed a Stochastic Gradient Descent (SGD) algorithm, which converges quicker than the vanilla SGD algorithm.

Some optimization has been done in the inference space as well. For example, Gemel~\cite{gemel} is a memory-efficient, real-time video analytics framework designed for edge computing that combines layers of models with similar architecture to make inference on edge more efficient. BGL~\cite{bgl} is another GPU-efficient GNN training framework that leverages a dynamic cache engine to reduce feature retrieving traffic.

\noindent \textbf{Uniqueness of \sol compared to prior work:} \sol is designed while keeping a specific goal in mind, which is to efficiently train multiple NN models simultaneously. This is particularly important when there are multiple users that need to train their NN models simultaneously, which is typically the case in existing cloud computing environments. To this end, the design of \sol employs several mechanisms to reduce memory consumption and training time while training multiple NN models simultaneously. As we present in Table~\ref{tab:frameworks}, none of the prior NN model training optimization frameworks is tailored towards this specific scenario, which is typical in today's cloud computing environments.


\begin{table*}[]
\centering
\caption{An overview of DNN training optimization frameworks.}
\label{tab:frameworks}
\begin{tabular}{|l|l|l|l|l|l|}
\hline
\textbf{Framework} & \textbf{Year} & \textbf{\makecell[c]{Strategy}}                                                  & \textbf{\makecell{Impact on\\accuracy}} & \textbf{\makecell{Compatibility with\\PyTorch/TensorFlow}} & \textbf{\makecell{Optimized for\\concurrent training}} \\ \hline
moDNN              & 2018          & \makecell[l]{Tune DNN training code to\\fit a memory budget}                      & Minimal                     & No    & No                                         \\ \hline
ZCOMP              & 2019          & \makecell[l]{ISA for fetching data in sequential\\blocks}                         & No                          & Yes (TensorFlow)  & No                            \\ \hline
Zico               & 2021          & \makecell[l]{Share memory resources across\\different training jobs}              & No                          & Yes (TensorFlow)     & Limited                          \\ \hline
PipeDream          & 2021          & \makecell[l]{Pipeline and distribute large models\\across multiple accelerators} & Minimal                     & Yes (PyTorch)     & No                             \\ \hline
DropIT             & 2022          & \makecell[l]{Eliminate cached tensor elements\\for better memory utilization}     & No                          & Yes (PyTorch)   & No                               \\ \hline
\textbf{\sol}            & \textbf{2024}          & \textbf{\makecell[l]{Combine multiple NN models to\\efficiently utilize GPU memory}}     & \textbf{No}                          & \textbf{Yes (PyTorch)}              & \textbf{Yes}                    \\ \hline
\end{tabular}
\end{table*}

\section{\sol Design} 
\label{sec:design}

In this section, we first present our design assumptions and an overview of the \sol design. We then discuss the components of the \sol design in detail. 


\subsection{Design Assumptions and Overview}

We assume a cloud or local environment where there are one or more servers available with a GPU cluster. This system is responsible for training NN models received via an online or local service. As it is typically the case with GPU clusters today, we assume that there are multiple user requests to train NN models. Servers use their CPUs to process the requests and their GPUs to train the models. 
\sol sits in the middle of the original Application Programming Interfaces (APIs) provided by the services. Cloud providers have the flexibility to use their own APIs and the frontends, and implement \sol in the backend. When there is a queue of NN models submitted from users for training, the framework accepts the model definitions of several models as input and combines them to produce a hybrid unified model based on the available resources. The hybrid model is then loaded to the acceleration hardware (most commonly GPUs) along with the datasets for the corresponding models in order to train the models based on the hyper-parameters provided by users. Once the hybrid model is trained, \sol separates each individual model and returns the trained models to the respective users.

\begin{figure}[t]
    \centering
    \includegraphics[width=\linewidth]{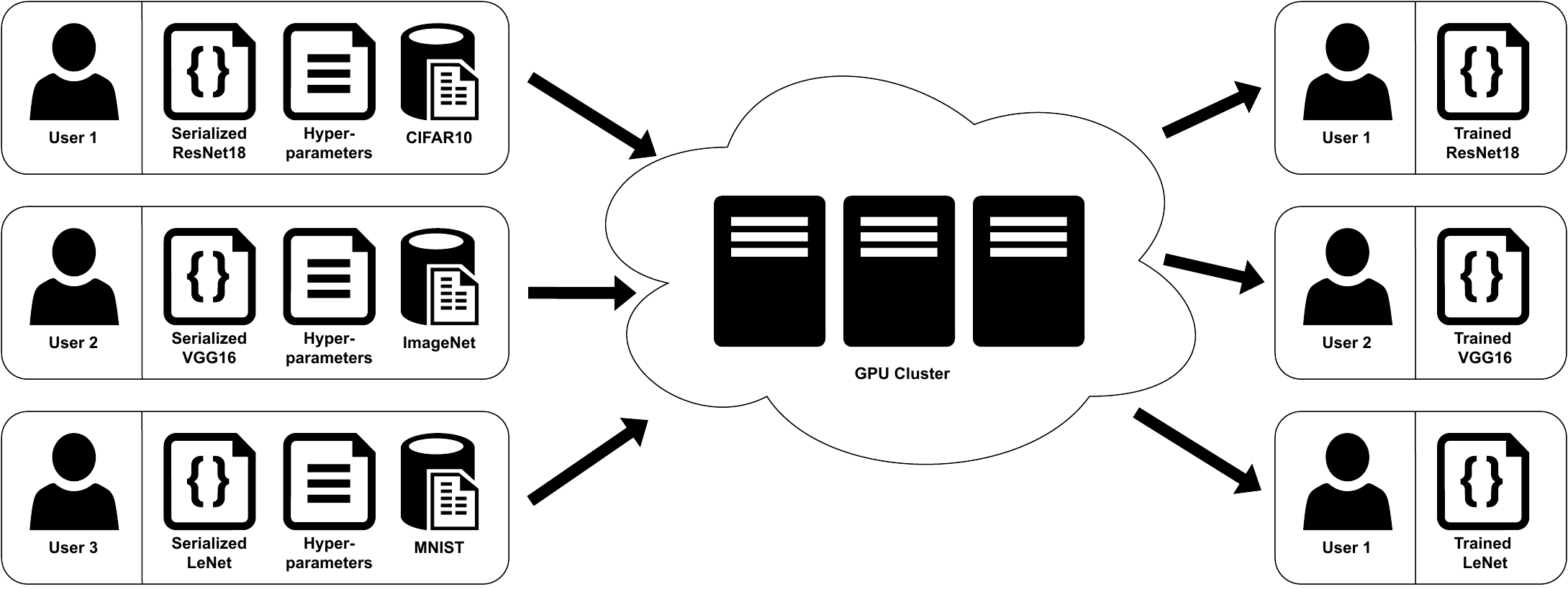}
    \includegraphics[width=\linewidth]{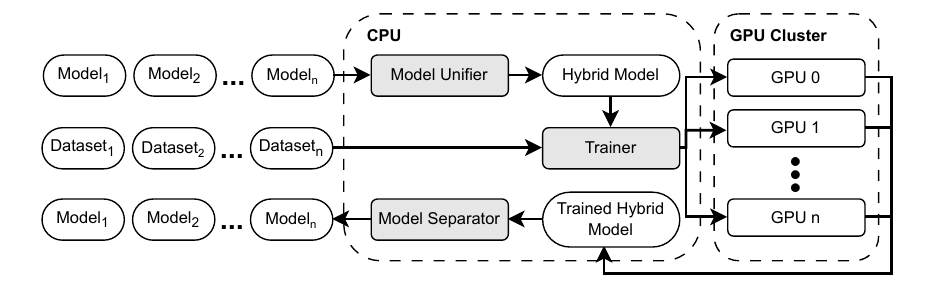}
    \caption{Workflow of the \sol operation.}
    \label{fig:workflow}
\end{figure}

The workflow of the \sol operation is presented in Figure~\ref{fig:workflow}. The framework accepts serialized model files containing the descriptions of the model architectures, the hyper-parameters, and the datasets (split into testsets and trainsets) in a \textit{tensor} format. It is up to the service provider to process the requests from the users and format the files accordingly. \sol reads the descriptions of the architectures from the serialized model files to generate the hybrid model. The system accepts serialized models in either \textit{.pt} or \textit{.pth} format and raw datasets in \textit{Tensor} format. The service also has options to accept hyperparameters, such as the learning rate, the batch size, the optimizer, and the schedulers in a \textit{json} file. Subsequently, the cloud service provider uses \sol to combine and train the user models. In this example, the service uses priority scheduling where User 2 has the highest priority, followed by User 1 and User 3 in Figure~\ref{fig:workflow}. The models will be trained in that order and the trained models will be returned to the users through the service.

Within the hybrid model architecture, each model definition is defined using a sequential module. As such, the hybrid model contains a sequential module for each sub-model. From here on, we will refer to an individual model inside the hybrid model as a ``sub-model''. \sol also generates different optimizers, criteria, and learning rate schedulers for each sub-model to satisfy the training requirements of each user. During a forward pass, the input data from the corresponding dataset is provided to the specific sub-model, which generates an output. Based on the output from the sub-model and the dataset, the framework then computes the gradient and loss using the specific criterion defined for that sub-model. Once the loss is computed, the backward pass is performed using the corresponding optimizer in order to adjust the weights of the particular sub-model. The same training procedure is applied to each sub-model training. Therefore, \sol can train multiple NN models by creating one large hybrid model without any impact on the training of the individual sub-models. \sol also offers different scheduling options for the service providers to decide the order of training the sub-models. \sol does not wait for all the sub-models to finish training. As soon as one sub-model is trained, a copy of the entire hybrid model is transferred to the CPU, while the training for the other sub-models continues in the GPU. Subsequently, \sol creates a model with the original model description from the user and copies the trained parameters from the particular sub-model into the original model description, which is ready to be delivered to the user. Since we assume that the CPU remains idle when the GPUs are busy training the models, the CPU can utilize the free time to create and separate a hybrid model. As such, \sol introduces no overhead on the overall performance of the service in the best-case scenario. There are three major components in \sol: (1) model unifier, (2) trainer, and (3) model separator. 

\subsection{Model Unifier}

\begin{figure*}[t]
    \centering
    \includegraphics[width=14cm]{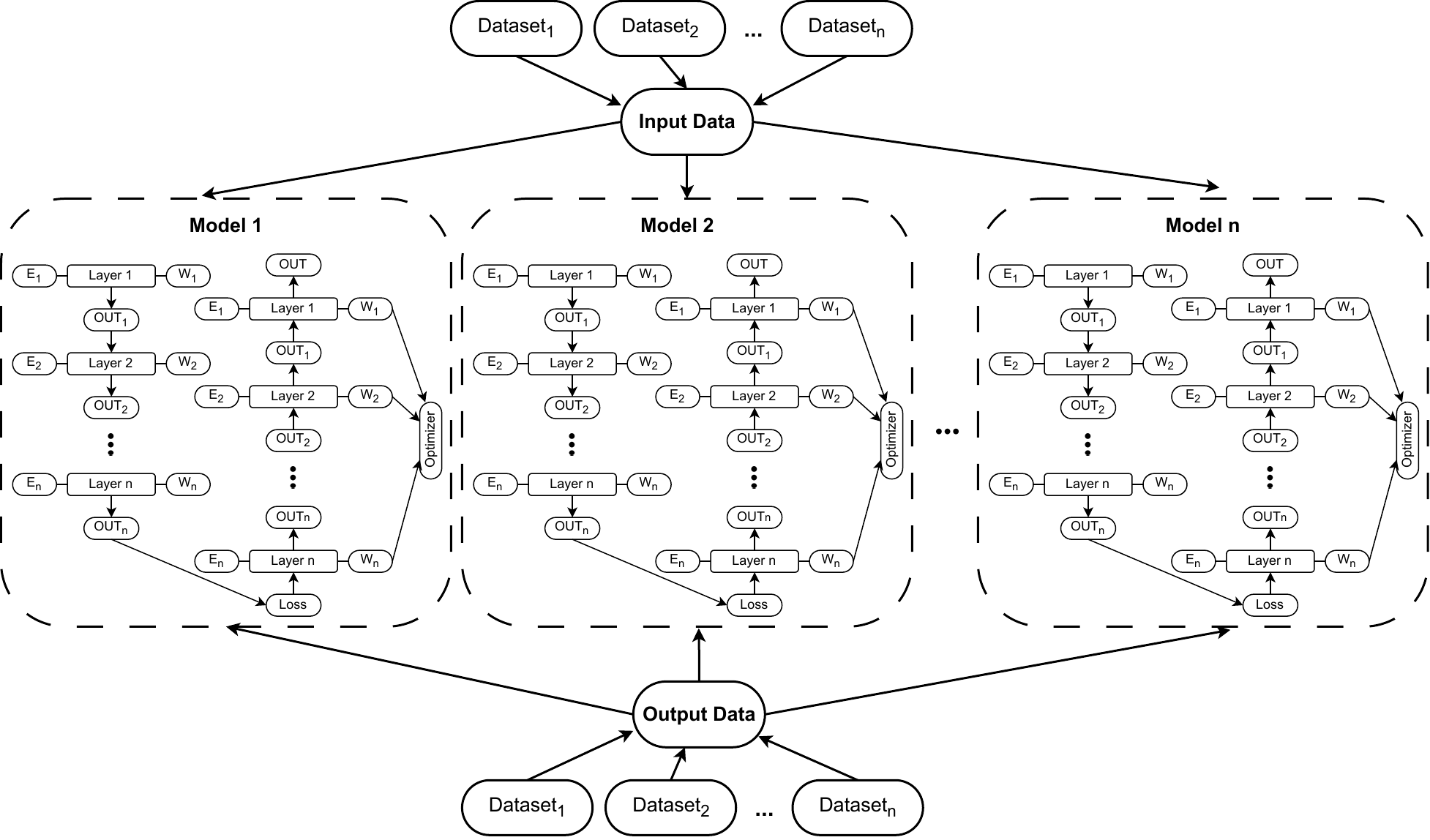}
    \caption{An example of a DAG of a hybrid model.}
    \label{fig:dag}
    \vspace{-0.2cm}
\end{figure*}

This component is responsible for merging multiple incoming models and generating the hybrid model that fits into the available GPU memory. \sol is designed to support any type of NN model architecture. As such, the model unifier can combine models of different architectures from different domains. For example, a computer vision model can be combined with an NLP model. The input consists of $n$ models ($M_1, M_2, M_3, \dotsc, M_n$) along with the hyper-parameters ($H_1, H_2, H_3, \dotsc, H_n$), and the optimizers ($O_1, O_2, \dotsc, O_n$) set for each sub-model. The component reads the definition of the models and creates the hybrid model $M_\text{hybrid}$ in a way that generates sub-graphs for each model within the hybrid model using Algorithm~\ref{algo:combine}. The final computation graph is a combination of all the Directed Acyclic Graphs (DAGs) from the $n$ models where $u_i^n$ is a layer (component), $\{u_i^n, u_j^n\}$ is the connection (edge) between two components, $H_k^n$ is the hyper-parameter, and $O^n$ is the optimizer of the model $n$. Each edge $\{u_i^n, u_j^n\}$ represents the input/output of each layer and $H_k^n$ is the set of hyper-parameters (\eg input size, batch size, learning rate). Figure~\ref{fig:dag} presents the DAG of the hybrid model containing $n$ different NN models. The hybrid graph $\mathcal{G}$ has one global input and one global output. Each sub-model contains two graphs for forward and backward propagation. A sub-model does not interfere with another, making sure that the training of one model does not impact the other. Moreover, each sub-model has its own optimizer based on the user-provided hyper-parameters. As such, when the weights are adjusted, it only affects the corresponding sub-model. According to the selected scheduling option, each batch from the provided datasets is fed into the model and the loss function is generated. Once the loss is generated, the back-propagation is calculated and the weights are adjusted according to the optimizer.


\begin{algorithm}
  \caption{\sol merging algorithm}
  \label{algo:combine}
  \renewcommand{\algorithmicrequire}{\textbf{Input:}}
  \renewcommand{\algorithmicensure}{\textbf{Output:}}
  \begin{algorithmic}[1]
  \Require Model architectures $\{m_1, m_2, \dotsc, m_n\}\in \mathcal{M}$, hyper-parameters $\{h_1, h_2, \dotsc, h_n\}$, Optimizers $\{o_1, o_2, \dotsc, o_n\}$, 

    \Ensure Merged $M_\text{hybrid}$

  \Statex \textit{Initialization} : $\mathcal{G} \leftarrow \emptyset$, $\mathcal{H} \leftarrow \emptyset$, $\mathcal{O} \leftarrow \emptyset$

  \For{$\forall m \in \mathcal{M}$}
    \For{$u_i, u_j \in m$}
        \State $\mathcal{G} \leftarrow \{u_i^m\}_{i=1}^n, \{u_i^m, u_j^m\}$
    \EndFor
    \State $\mathcal{H} \leftarrow h_m^k$
    \State $\mathcal{O} \leftarrow o_m$
   \EndFor
   \State $M_\text{combined} \leftarrow \langle \mathcal{G}, \mathcal{H}, \mathcal{O} \rangle$\\
   \Return $M_\text{hybrid}$
\end{algorithmic}
\end{algorithm}

\subsection{Trainer}

The trainer component sets the hybrid model for training. Once the hybrid model is created by the model unifier component, the hybrid model is sent to the trainer component. Additionally, this component accepts all batched datasets $\{b_1, b_2, \dotsc, b_n\}$ along with the hyper-parameters needed to train the models. The trainer component then assigns the training job to a GPU available in the GPU cluster. The gradient and loss are calculated for each individual sub-model with its own optimizers using Algorithm~\ref{algo:train}. The cloud provider can select a scheduler from the available scheduling options. Currently, \sol offers four scheduling options available for a provider to choose from: First Come First Serve, Priority Scheduling, Shortest Job First, and Round Robin.

\begin{algorithm}
  \caption{\sol hybrid graph training algorithm}\label{algo}
  \label{algo:train}
  \renewcommand{\algorithmicrequire}{\textbf{Input:}}
  \renewcommand{\algorithmicensure}{\textbf{Output:}}
  \begin{algorithmic}[1]
  \Require Hybrid model $M_\text{hybrid}$, batched datasets $\{b_1, b_2, \dotsc, b_n\}$ 
  \Ensure Trained $\theta^1_{b_1}, \theta^2_{b_2}, \dotsc, \theta^m_{b_m}$

  \Statex \textit{Initialization} : Random $\theta^1_0, \theta^2_0, \dotsc, \theta^m_0$

  \For{$\forall d, \in b_i$}
        \State $g^m_{b_i}(d_i) \leftarrow \nabla_{\theta^m_d}\mathcal{L}(\theta^m_d, d_i') \text{ for } \forall m \in M_\text{combined}$
        \State $\theta^m_{b_i + 1} \leftarrow \theta^m_{b_i} - \eta_{b_i} g^m_{b_i}$
   \EndFor
   \Return Trained $\theta^1_{b_1}, \theta^2_{b_2}, \dotsc, \theta^m_{b_m}$
\end{algorithmic}
\end{algorithm}

\noindent \textbf{First Come First Serve:} First Come First Serve is recommended if a cloud provider would like to serve users in the order of their arrival. The first sub-model within the hybrid model will be trained first. Once the training finishes, the first trained model is returned to the user while the second sub-model in the queue is being trained. Subsequently, all the models in the queue are trained sequentially and returned back to the users.

\noindent \textbf{Priority Scheduling:} Priority Scheduling is useful when a cloud provider would like to set priorities for specific jobs from the queue. Each sub-model within the hybrid model is assigned a priority value (a lower value indicates a higher priority). The sub-models are trained according to the priorities and returned to the corresponding users as soon as a sub-model finishes training. If two or more models have the same priority, a round-robin approach will be applied among them.

\noindent \textbf{Shortest Job First:} A cloud provider has the option to train the smallest sub-model first. This is an efficient way to utilize memory as the entire memory will not be reserved first as our evaluation results demonstrate in Section~\ref{sec:eval}. \sol currently supports two different ways of defining the length of a training job: it can either be the size of the model or the number of epochs that a model needs to be trained. 

\noindent \textbf{Round Robin:} Round Robin is another scheduling algorithm available in \sol. This algorithm allows all the sub-models to train concurrently. Sub-models are trained in turn after one sub-model finishes one epoch. This approach is currently non-preemptive; as such, smaller sub-models will suffer if there is a large disparity in terms of the time taken for a single epoch by each sub-model. Therefore, this scheduling algorithm is recommended for hybrid models consisting of sub-models that require approximately the same amount of time to be trained per epoch.

Other advanced scheduling algorithms can be implemented by the cloud providers in order to satisfy their specific needs. Once a scheduling algorithm is selected, the trainer then assigns the hybrid model to one of the available GPUs in the GPU cluster. If the GPU cluster is heterogeneous, hybrid model training is assigned to a GPU with available memory. The memory requirement for a hybrid model can be calculated as follows:

\begin{equation}
    mem(M_i) = mem_{u}(M_i) + mem_{r}(M_i) + mem_{CUDA}(M_i)
\end{equation}
\begin{equation}
    mem(M_h) = max\{mem(M_i) | i \in n\},
\end{equation}

where $mem(M_i)$ is the total memory consumed by a sub-model $M_i$, $mem_{u}(M_i)$ is the unreleased memory, $mem_{r}(M_i)$ is the reserved memory, and $mem_{CUDA}(M_i)$ is the memory required for the CUDA context. The final memory estimation for the hybrid model $M_h$ is $mem(M_h)$.

\subsection{Model Separator}

This component is responsible for the extraction of one or more trained models from the hybrid model. When the model definitions are provided to \sol, it stores the original model definition $M$ for each sub-model from the hybrid model $M_\text{hybrid}$. Once the training job of a sub-model is finished, the trainer component provides a copy of the entire hybrid model $M_\text{hybrid}$ to the model separator component, while the other sub-models are being trained. The model separator component then regenerates the original model $M_\text{original}$ and the model definition $M$, and copies the trained parameters $\{w_1, w_2, w_3, \dotsc, w_n\}$ of that specific sub-model to the original model $M_\text{original}$. The trained model is then ready to be returned to the user. In other words, \sol returns each trained sub-model to the corresponding user as soon as the training of this specific sub-model is done without having to wait for the training of all other sub-models to finish. This is particularly useful when heterogeneous sub-model architectures (\eg sub-models of substantially different sizes or sub-models that require vastly different times per epoch during training) are combined and acts complementary to the selection of an appropriate scheduling algorithm. As a result, sub-models that finish training first do not need to wait for sub-models that take a longer time to train.

We designed \sol to support pausing and resuming training of a particular sub-model without impacting the rest of the hybrid model (rest of the sub-models). Users can create checkpoints at any given time and resume training from a checkpoint in the future. When requested, \sol stores the states from that particular sub-model and the corresponding optimizer, which can be provided to the user.

\subsection{Memory Conservation and Time-Saving Mechanisms}

A fundamental goal of \sol is to save on the available GPU memory, thus allowing a cloud provider to scale NN model training (train more models) at the same time. To achieve that, \sol employs several memory conservation mechanisms. Additionally, as an outcome of these mechanisms, \sol can, in general, reduce training time as well. These mechanisms range from optimizing memory allocations to reducing the number of I/O operations.

NN training frameworks typically execute mathematical operations on trainable weights (operators) on both the forward and the backward pass. To compute these mathematical operations, certain library information and temporary tensors are required, which are allocated in the GPU memory. Once the computation is completed, the GPU memory is not immediately available for others to use. When an NN model is trained on a GPU, the total memory allocation can be classified into four dimensions~\cite{dnnmem}. We discuss the impact of these dimensions and how \sol capitalizes on them to reduce GPU memory consumption below.

\noindent \textbf{Weight tensors:} In an NN model, there is a number of learnable parameters and an equal number of weight gradients. Weights (including weight biases) are used during forward pass operations and the gradients are used to adjust the weights during backward pass operations. Both weight parameters and gradients sit in the GPU memory. As \sol does not modify the model architectures, GPU memory consumption remains the same as in training the models concurrently.

\noindent \textbf{Input/output tensors:} During NN model training, the input data is fed into the model in the form of mini-batches. The larger the mini-batch size is, the more GPU memory it consumes. Since the size of mini-batches is a hyper-parameter, it will be set by the user. This is where \sol can save a substantial amount of memory. When multiple models are trained concurrently, input mini-batches of all models are loaded at once; however, \sol loads them on-demand according to the scheduling algorithms discussed earlier. Operations of output calculations and gradients also happen according to the corresponding scheduling scheme, which further reduces GPU memory consumption. Moreover, if multiple models need to be trained using a common dataset, the common dataset is loaded only once into the memory, resulting in further memory conservation.

\noindent \textbf{Ephemeral tensors:} Libraries and APIs needed for training consume additional GPU memory. 
For example, cuDNN APIs~\cite{cudnn} are required each time a model is loaded into a GPU. As such, the memory consumption increases linearly with the number of models to be trained. Since \sol generates a single hybrid model, these libraries and APIs are loaded only once, conserving GPU memory. Moreover, some temporary tensors allocate GPU memory, which helps in faster operation. Since \sol offers more free GPU memory space, sometimes it helps in faster operation.

\noindent \textbf{Resident buffer:} CUDA context contains information required to manage an NVIDIA GPU, which is loaded with each model. As \sol creates a single hybrid model, the CUDA context is also loaded only once. Internal tensor fragmentation also requires additional memory space, which is reduced through the use of a single hybrid model. Finally, miscellaneous reservations like fusion buffers are also reduced through the use of a single hybrid model.

\section{Evaluation}
\label{sec:eval}

Using our \sol prototype, we evaluated \sol in the following scenarios: (1) we compared \sol to vanilla PyTorch; (2) we evaluated its performance for different scheduling options; and (3) we compared \sol to two state-of-the-art memory optimization frameworks for NN model training. Our evaluation metrics include the overall memory consumption during training, the training time (time for each individual model to finish training and the overall time for multiple models to finish training as a whole), and the training convergence of the NN models.

\subsection{Evaluation setup}

We selected six different models and datasets to cover a wide range of real-world applications and use cases: from small to large convolutional neural network models for computer vision applications to different natural language processing models. We conducted our experiments on a system with an AMD Ryzen Threadripper PRO 5955WX (16-Cores), 128 GB of system memory, and an Nvidia GeForce RTX 4090 GPU (24 GB). Table~\ref{tab:eval_models} presents the models, datasets, and the hyper-parameters used for our evaluation. For every model, the memory consumption in MiB, elapsed time in seconds, and the training/testing accuracy were recorded.

\begin{table*}[]
\centering
\caption{Evaluation models, datasets, and hyper-parameters}
\label{tab:eval_models}
\small
\begin{tabular}{|l|l|l|l|l|l|l|}
\hline
\textbf{Model}  &\textbf{No. of Parameters}             & \textbf{Dataset}    & \textbf{Epochs} & \textbf{Batch Size} & \textbf{Learning Rate} & \textbf{Optimizer}\\ \hline
LeNet     & $44470$        & MNIST     & 20     & 256        & 0.01   & SGD       \\ \hline
ResNet18   & $11.17\times10^6$       & CIFAR10    & 100    & 128        & 0.001     & SGD    \\ \hline
ResNet50   & $21.32\times10^6$      & CIFAR100   & 120    & 128         & 0.001     & SGD    \\ \hline
VGG16 (Fine-tune) & $13.83\times10^7$ & Imagennete & 10     & 64         & 0.002  & SGD       \\ \hline
Transformer (NLP) & $12.02\times10^6$ & WikiText   & 10    & 10         & 5.0     & SGD     \\ \hline
XLM-RoBERTa (NLP) & $27.80\times10^8$ & SST-2     & 5     & 16        & 0.00001     & Adam     \\ \hline
\end{tabular}
\end{table*}

We discuss the NN models, the datasets, and our rationale for choosing each model to represent real-world use cases below.

\noindent \textbf{LeNet:} LeNet~\cite{lenet} is a simple convolutional neural network model with about 44 thousand trainable parameters. This model is trained with the MNIST~\cite{mnist} dataset. We selected this model to represent the impact of a small model when it is combined with other models. The MNIST dataset contains 60,000 single-channel images of handwritten numbers. We trained this model for 20 epochs with a batch size of 256 and a learning rate of 0.01.

\noindent \textbf{ResNet18:} ResNet~\cite{resnet} is a popular image classification model that contains residual blocks and skip connections. This model represents the impact of a medium-sized model in the operation of \sol. We trained ResNet18 with the CIFAR10~\cite{cifar} dataset. The ResNet18 model has more than 11 million trainable parameters. The CIFAR10 dataset contains 60,000 images of 10 classes of different objects. We trained this model for 100 epochs with a scheduler for 100 epochs with a learning rate of 0.001 and a batch size of 128.

\noindent \textbf{ResNet50:} ResNet50 model represents the impact of a somewhat large-sized model in the operation of \sol, which has more than 21 million trainable parameters. We trained ResNet50 with the CIFAR100~\cite{cifar} dataset. The CIFAR10 dataset contains 60,000 images of 100 classes of different objects. We trained this model with a scheduler for 120 epochs with a learning rate of 0.002 and a batch size of 64.

\noindent \textbf{Custom VGG16:} VGG~\cite{vgg} is another popular image classification model that contains multiple blocks of convolutional, ReLU, and max-pooling layers. Here, we have taken a pre-trained VGG16 model, which was trained on the ImageNet~\cite{imagenet} dataset. The VGG16 model is modified by inserting CBAM~\cite{cbam} attention modules between each VGG block. This custom VGG16 model has more than 138 million trainable parameters. Finally, we trained this custom model with the Imagenette dataset~\cite{imagenette}. Imagenette is a subset of the full ImageNet dataset containing 10 classes. This model demonstrates an application of transfer learning, which is a widely-used approach today.

\noindent \textbf{Transformer model:} Transformer models are one of the most popular approaches when it comes to language models nowadays~\cite{Transformer}. Large language models like GPT-4 are powered by transformer models~\cite{gpt4}. The transformer model has more than 12 million trainable parameters. We trained a transformer model with the WikiText dataset~\cite{wikitext} with a learning rate of 5.0, a batch size of 10, and 10 epochs. This model represents the effect of language models.

\noindent \textbf{XLM-RoBERTa:} RoBERTa is a popular replication study of a popular language model known as BERT~\cite{roberta}. XLM-RoBERTa is a pre-trained multilingual model that outperforms BERT~\cite{xlm}. The XLM-RoBERTa has more than 278 million trainable parameters. We trained XLM-RoBERTa using the SST-2 dataset~\cite{SST} for 5 epochs, a batch size of 16, and a learning rate of 1e-5. This model presents another example of NLP training.


\subsection{Comparison with vanilla PyTorch}

We compare \sol with vanilla PyTorch to investigate its design and performance in comparison with a baseline. For this comparison, we used PyTorch version 2.0 with CUDA 11.7, which are the latest versions available at the time of this experiment. We trained multiple models from Table~\ref{tab:eval_models} concurrently and sequentially to compare with \sol based on training time, accuracy, and GPU memory usage. The reason for comparing the accuracy of models trained through \sol with the accuracy of models trained through PyTorch is to demonstrate that there is no impact on the training/testing accuracy of the models when trained through \sol. We first selected three models from Table~\ref{tab:eval_models} (LeNet, ResNet18, and ResNet50) and trained them with \sol. We then selected the remaining three models from Table~\ref{tab:eval_models} (VGG16, Transformer, and XLM-RoBERTa) and trained them with \sol.

\begin{figure}
     \centering
     \begin{subfigure}[b]{0.48\linewidth}
         \centering
         \includegraphics[width=\textwidth]{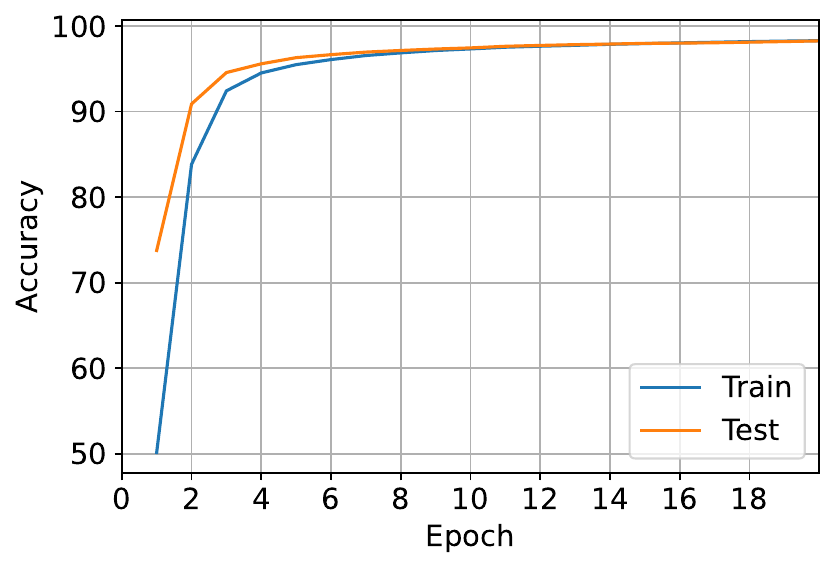}
         \label{fig:lenet_train}
         \caption{Vanilla PyTorch}
     \end{subfigure}
          \begin{subfigure}[b]{0.48\linewidth}
         \centering
         \includegraphics[width=\textwidth]{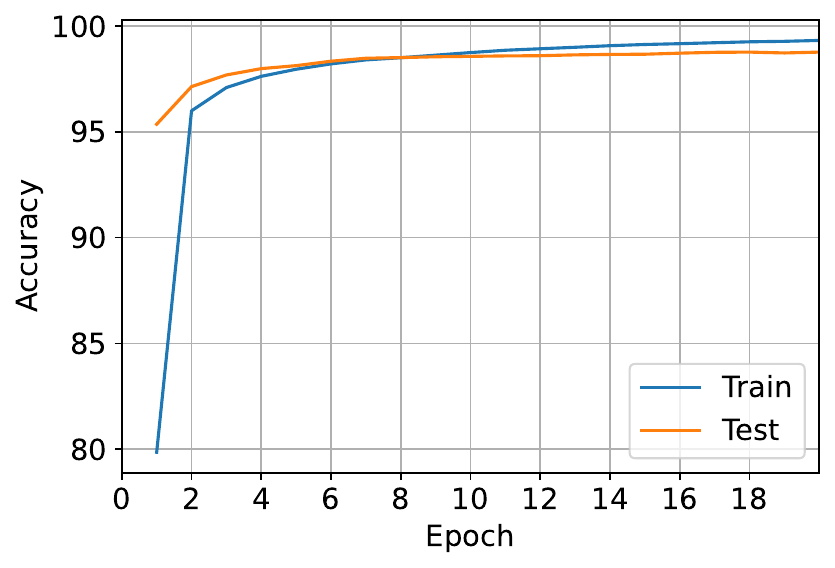}
         \label{fig:lenet_hybrid_train}
         \caption{\sol}
     \end{subfigure}
        \caption{Comparison of LeNet training with vanilla PyTorch and \sol.}
        \label{fig:lenet_train_comp}
\end{figure}

\begin{figure}
     \centering
     \begin{subfigure}[b]{0.48\linewidth}
         \centering
         \includegraphics[width=\textwidth]{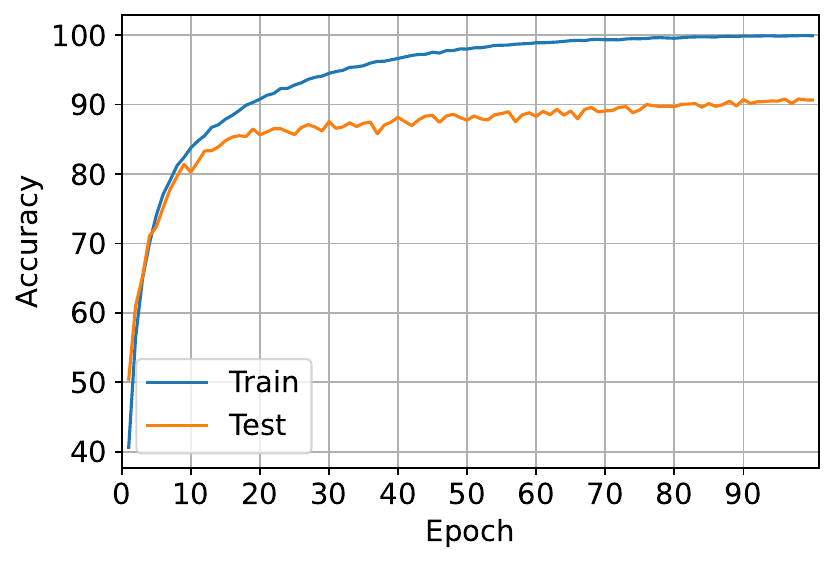}
         \label{fig:resnet18_train}
         \caption{Vanilla PyTorch}
     \end{subfigure}
          \begin{subfigure}[b]{0.48\linewidth}
         \centering
         \includegraphics[width=\textwidth]{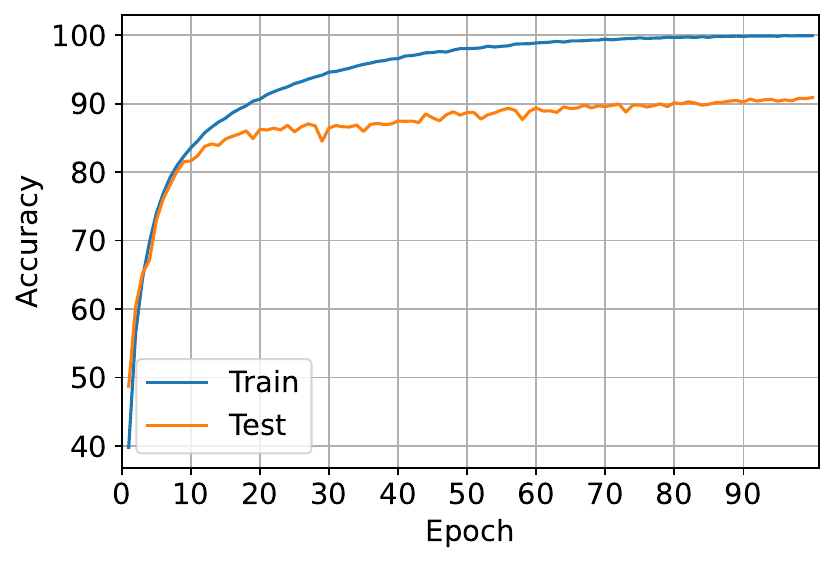}
         \label{fig:resnet18_hybrid_train}
         \caption{\sol}
     \end{subfigure}
        \caption{Comparison of ResNet18 training with vanilla PyTorch and \sol.}
        \label{fig:resnet18_train_comp}
\end{figure}

\begin{figure}
     \centering
     \begin{subfigure}[b]{0.48\linewidth}
         \centering
         \includegraphics[width=\textwidth]{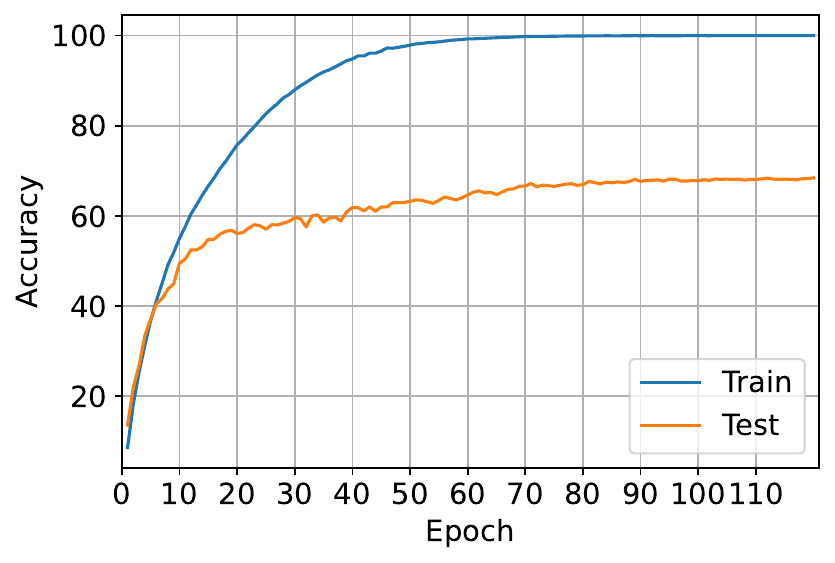}
         \label{fig:resnet50_train}
         \caption{Vanilla PyTorch}
     \end{subfigure}
          \begin{subfigure}[b]{0.48\linewidth}
         \centering
         \includegraphics[width=\textwidth]{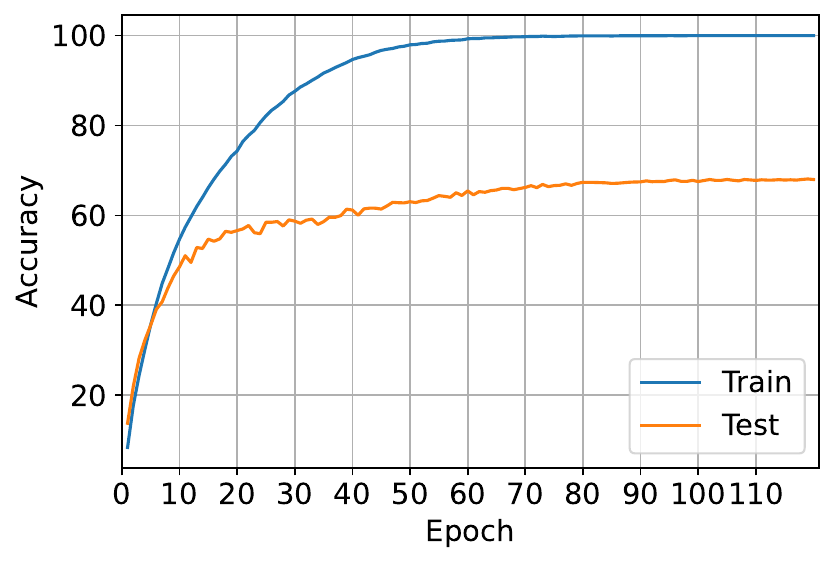}
         \label{fig:resnet50_hybrid_train}
         \caption{\sol}
     \end{subfigure}
        \caption{Comparison of ResNet50 training with vanilla PyTorch and \sol.}
        \label{fig:resnet50_train_comp}
\end{figure}

\noindent \textbf{LeNet, ResNet18, and ResNet50:} First, we trained LeNet, ResNet18, and ResNet50 concurrently with vanilla PyTorch. Next, we combined LeNet, ResNet18, and ResNet50 and trained them with \sol. Figure~\ref{fig:lenet_train_comp} presents the training and testing accuracy comparison of the LeNet model when trained with the MNIST dataset through vanilla PyTorch and \sol for 20 epochs. Our results show that \sol did not impact the training of the LeNet model. With vanilla PyTorch, the training and the validation accuracy were 99.3\% and 98.77\% respectively; in comparison, they were 99.3\% and 98.78\% when trained with \sol. Figure~\ref{fig:resnet18_train_comp} presents the training and testing accuracy comparison of the ResNet18 model when trained with CIFAR10 for 100 epochs. Our results show that \sol did not impact the training of the ResNet18 as well. The training and the validation accuracy with vanilla PyTorch were 99.9\% and 90.8\% respectively; whereas, \sol achieved 99.5\% and 90.5\% accuracy for the training and validation accuracy respectively. Finally, the training and testing accuracy comparison of the ResNet50 model when trained with CIFAR100 for 120 epochs is presented in Figure~\ref{fig:resnet50_train_comp}, which again indicate no impact on accuracy when the model is trained with \sol. The training accuracy with vanilla PyTorch was 99.98\%, and the validation accuracy was 68.1\%. \sol achieved a training accuracy of 99.98\% and a validation accuracy of 68.0\%. These results show that \sol can train multiple models without impacting the training process of individual models.

\begin{figure}
     \centering
     \begin{subfigure}[b]{0.48\linewidth}
         \centering
         \includegraphics[width=\textwidth]{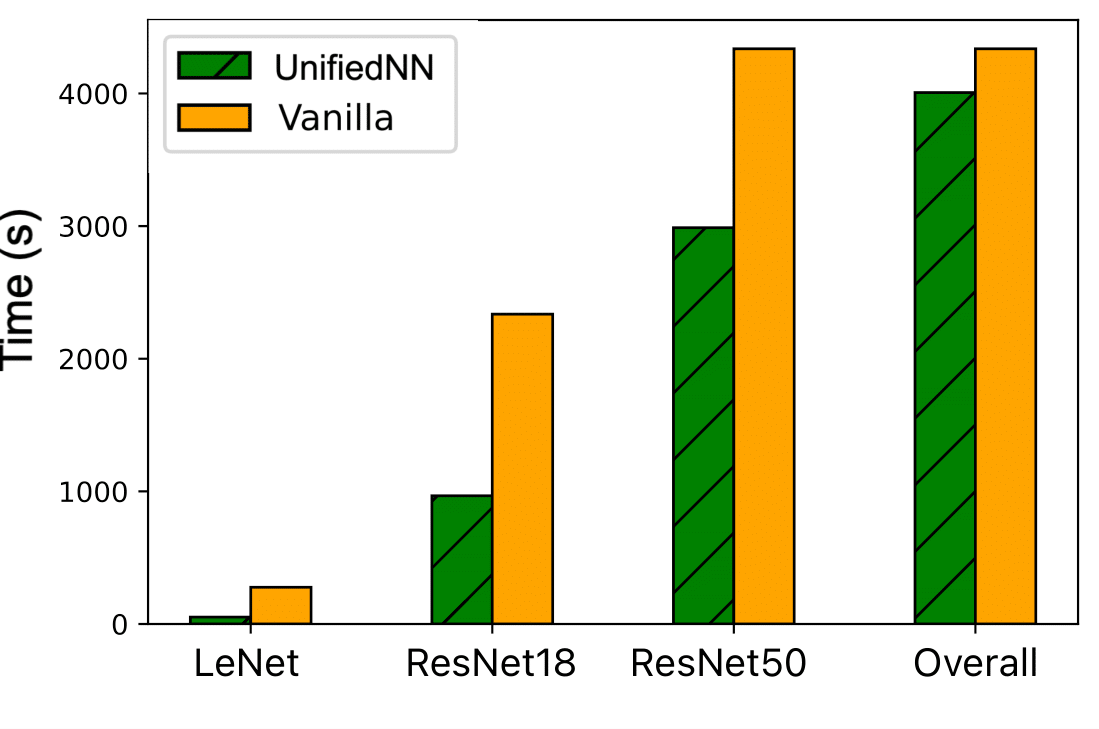}
         \caption{Training time}
     \end{subfigure}
          \begin{subfigure}[b]{0.48\linewidth}
         \centering
         \includegraphics[width=\textwidth]{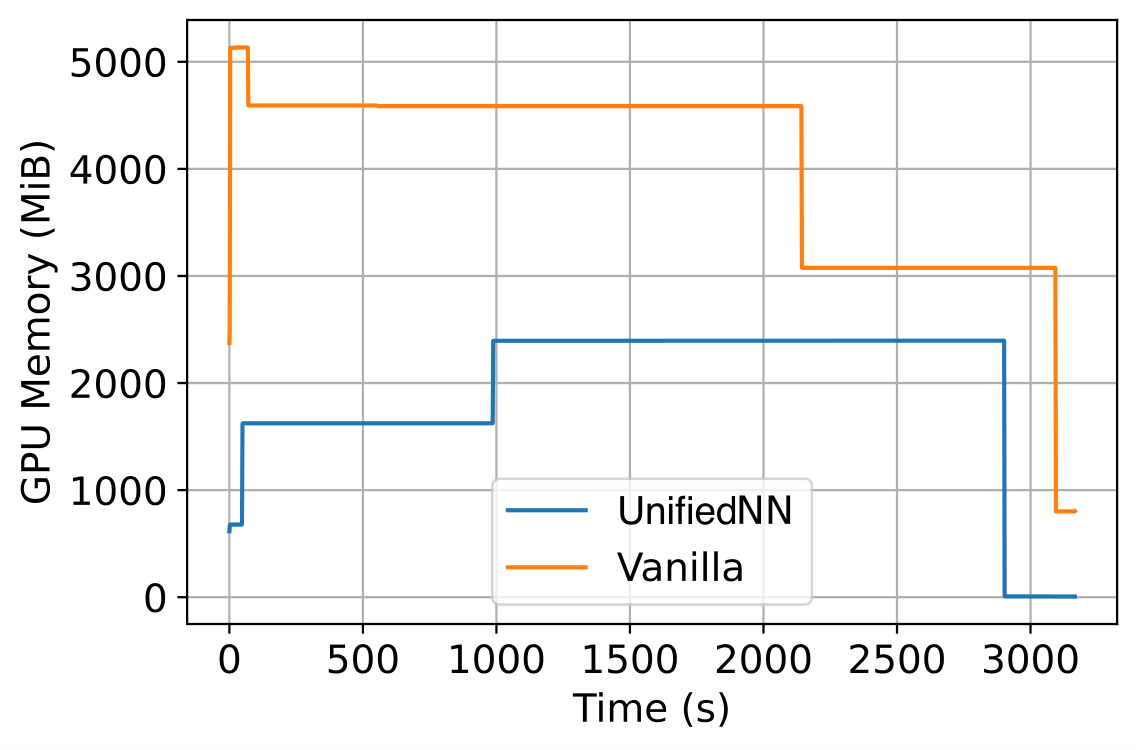}
         \caption{GPU memory usage}
     \end{subfigure}
        \caption{Comparison training time and GPU memory usage of LeNet, ResNet18, and ResNet50 when trained through vanilla PyTorch and \sol.}
        \label{fig:lenet_resnet_time}
\end{figure}

Next, we compare the training time for each model. We concurrently trained LeNet, ResNet18, and ResNet50 with vanilla PyTorch and recorded the total training time for each model. For this comparison, we selected the priority scheduling where LeNet had the highest priority, followed by ResNet18 and ResNet50, respectively. Figure~\ref{fig:lenet_resnet_time} presents the training times of the models with different approaches for an average of three runs. When trained concurrently with vanilla PyTorch, LeNet took four and a half minutes, ResNet18 took about 39 minutes, and ResNet50 took about 56 minutes on average. When trained through \sol, the training process for LeNet took less than a minute, about 16 minutes for ResNet18, and about 50 minutes for ResNet50 on average, which are improvements of 81\%, 58\%, and 13\%, respectively, compared to vanilla PyTorch. Overall, \sol was 7.6\% faster than vanilla PyTorch.

Finally, we compared the GPU memory usage while the models were being trained with vanilla PyTorch and \sol. Figure~\ref{fig:lenet_resnet_time} presents the GPU memory consumption of both vanilla PyTorch and \sol. At the beginning, we observe a spike in GPU memory consumption for vanilla PyTorch, since the models were trained concurrently, thus all models and necessary parameters were loaded into the memory at once. When a model finished training, the occupied memory was released. As such, memory usage decreased over time. In comparison, \sol occupies memory more efficiently. Since the LeNet model was trained first, it did not require as much memory. Although there is a peak during the training of larger models, it is not nearly as much as for vanilla PyTorch. The peak GPU memory consumption for vanilla PyTorch was 5134 MiB on average. On the other hand, \sol peaked at 2396 MiB on average for three runs, therefore, reducing the overall memory consumption by 53.3\%.

\begin{figure}
     \centering
     \begin{subfigure}[b]{0.48\linewidth}
         \centering
         \includegraphics[width=\textwidth]{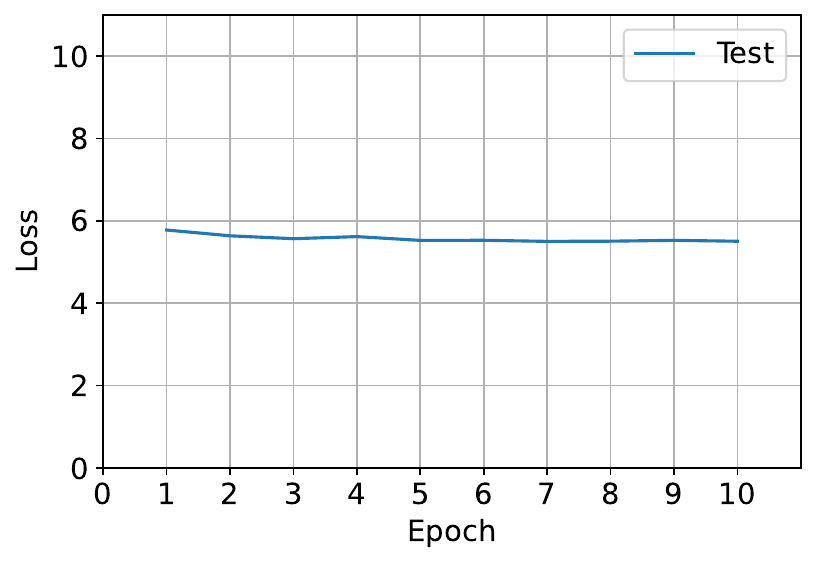}
         \label{fig:trans_train}
         \caption{Vanilla PyTorch}
     \end{subfigure}
          \begin{subfigure}[b]{0.48\linewidth}
         \centering
         \includegraphics[width=\textwidth]{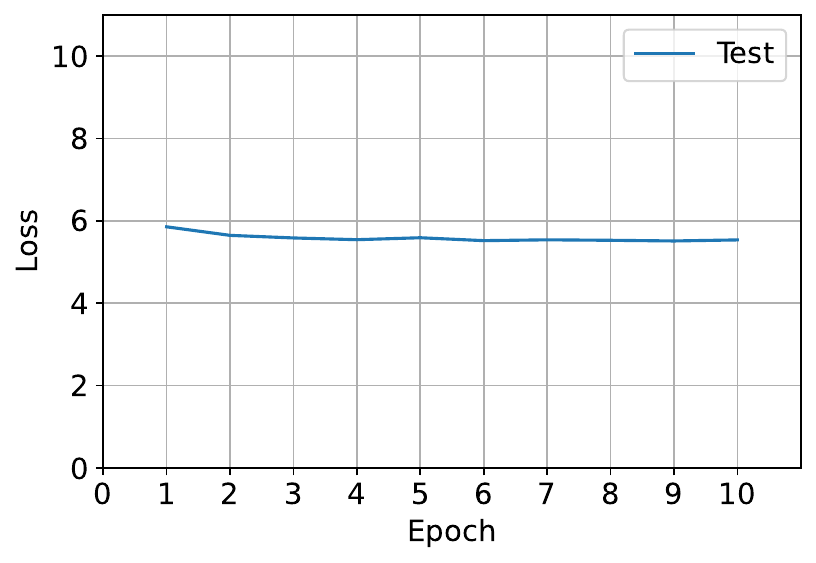}
         \label{fig:trans_hybrid_train}
         \caption{\sol}
     \end{subfigure}
        \caption{Comparison of the transformer model training with vanilla PyTorch and \sol.}
        \label{fig:trans_train_comp}
\end{figure}
\begin{figure}
     \centering
     \begin{subfigure}[b]{0.48\linewidth}
         \centering
         \includegraphics[width=\textwidth]{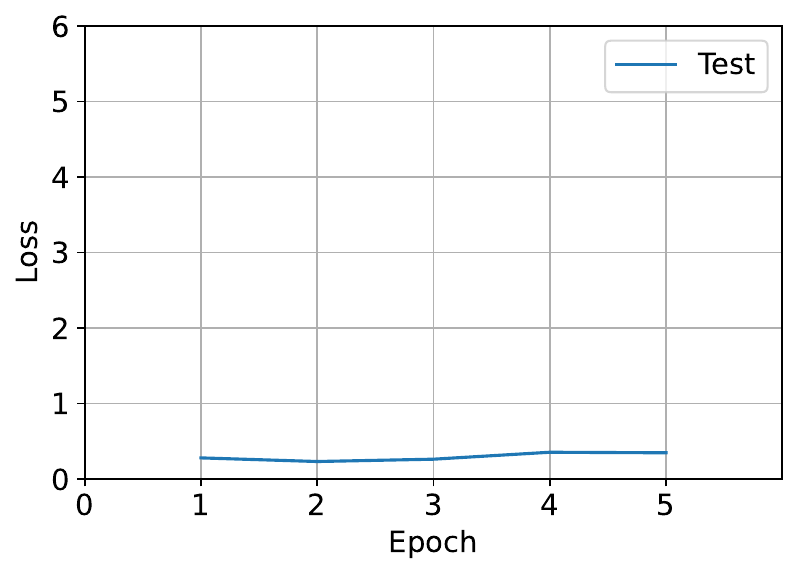}
         \label{fig:roberta_train}
         \caption{Vanilla PyTorch}
     \end{subfigure}
          \begin{subfigure}[b]{0.48\linewidth}
         \centering
         \includegraphics[width=\textwidth]{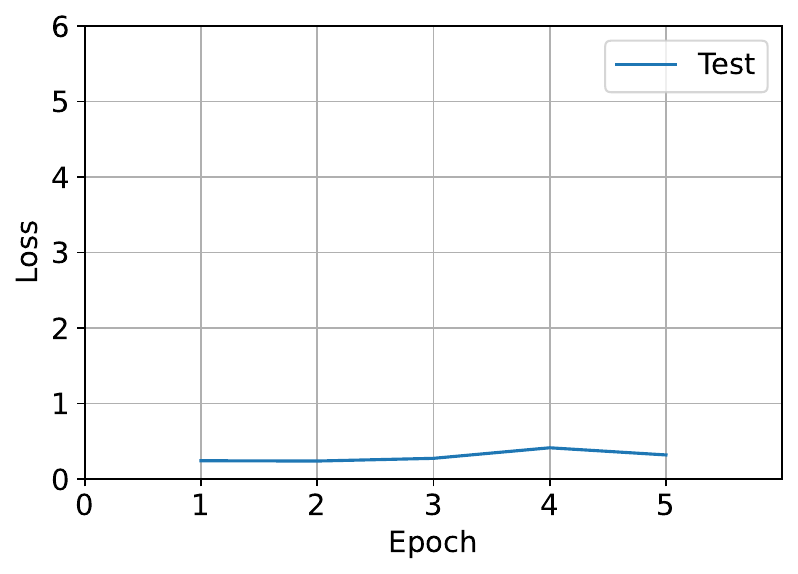}
         \label{fig:roberta_hybrid_train}
         \caption{\sol}
     \end{subfigure}
        \caption{Comparison of XLM-RoBERTa training with vanilla PyTorch and \sol.}
        \label{fig:roberta_train_comp}
\end{figure}

\noindent \textbf{VGG16, Transformer, and XLM-RoBERTa:} Next, we trained the custom VGG16, a transformer model, and XLM-RoBERTa concurrently with vanilla PyTorch and \sol to compare the training convergence, training time, and the GPU memory consumption. 
Priority scheduling was also used in this case as well. The highest priority was given to the transformer model, followed by XLM-RoBERTa and the custom VGG16, respectively. Figure~\ref{fig:trans_train_comp} presents the validation loss of the transformer model and Figure~\ref{fig:roberta_train_comp} presents the validation loss of the XLM-RoBERTa model. As the results suggest, the training process of the models was not impacted by \sol in this case as well. The loss convergences for both models were mostly the same to vanilla PyTorch. The transformer model achieved a loss of 4.75 during training and 5.51 during testing using vanilla PyTorch. Using \sol, it achieved a loss of 4.75 during training and 5.50 during testing. The same applies to XLM-RoBERTa. The validation loss and accuracy with vanilla PyTorch were 0.34 and 91.62\% respectively; whereas, the validation loss and accuracy with \sol were 0.35 and 91.74\%, indicating that \sol indeed had no impact on accuracy. 


\begin{figure}[t]
     \centering
     \begin{subfigure}[b]{0.48\linewidth}
         \centering
         \includegraphics[width=\textwidth]{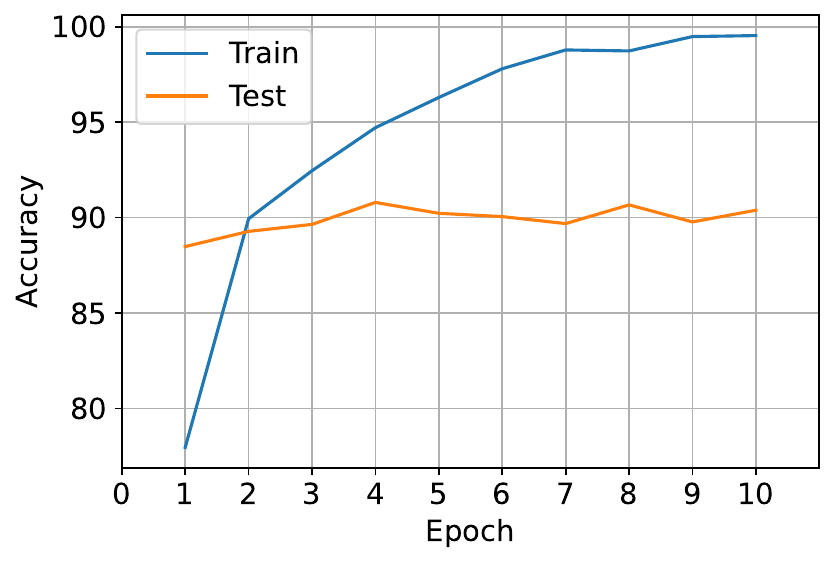}
         \label{fig:vgg_train}
         \caption{Vanilla PyTorch}
     \end{subfigure}
          \begin{subfigure}[b]{0.48\linewidth}
         \centering
         \includegraphics[width=\textwidth]{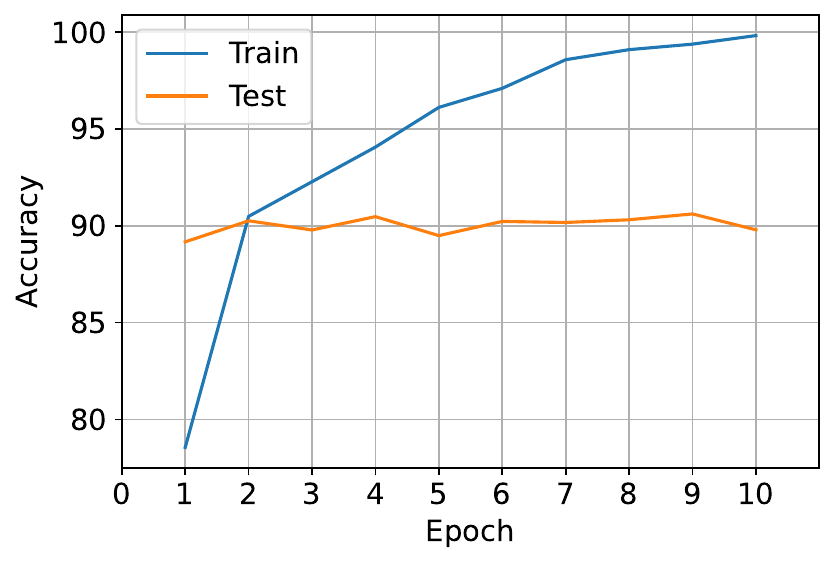}
         \label{fig:vgg_hybrid_train}
         \caption{\sol}
     \end{subfigure}
        \caption{Comparison of VGG16 training with vanilla PyTorch and \sol.}
        \label{fig:vgg_train_comp}
\end{figure}

Figure~\ref{fig:vgg_train_comp} presents the results of the training process for the custom VGG16 model. After fine-tuning the model with the Imagenette dataset for 10 epochs, the training and the testing accuracy with vanilla PyTorch for this model were 99.53\% and 90.38\%; whereas, they were 99.61\% and 90.44\%, respectively with \sol. \sol did not interfere with the training process in this case as well.

In terms of training time, the training of the transformer model took slightly less than 8 minutes, the training of XLM-RoBERTa took about 23 minutes, and the training of the custom VGG16 took 10 and a half minutes on average with vanilla PyTorch. On the other hand, the training of the transformer model took less than 2 minutes, the training of XLM-RoBERTa took 13 minutes, and the training of the custom VGG16 took about 4 and a half minutes on average when trained through \sol. These results demonstrate that \sol substantially reduces training time when compared with vanilla PyTorch. Specifically, the training time improvements through \sol for the transformer model, XLM-RoBERTa, and the custom VGG16 model are 76\%, 44\%, and 56\%, respectively. Overall, \sol was 16\% faster than vanilla PyTorch.

\begin{figure}
     \centering
     \begin{subfigure}[b]{0.48\linewidth}
         \centering
         \includegraphics[width=\textwidth]{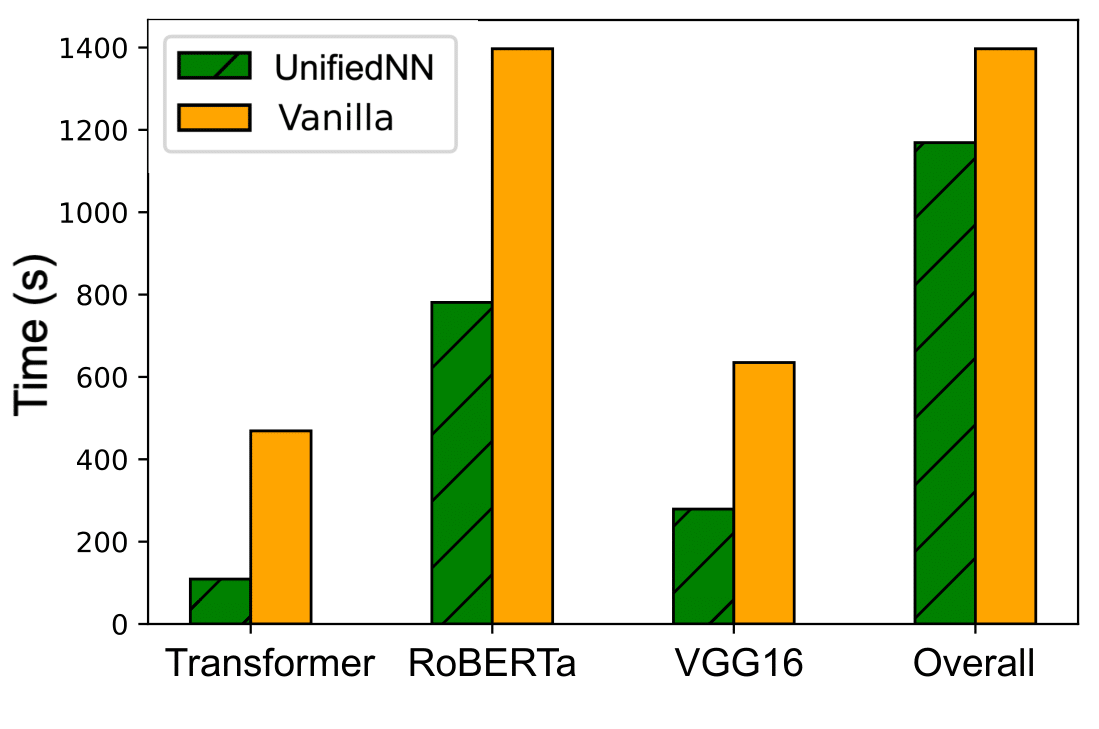}
         \caption{Training time}
     \end{subfigure}
          \begin{subfigure}[b]{0.48\linewidth}
         \centering
         \includegraphics[width=\textwidth]{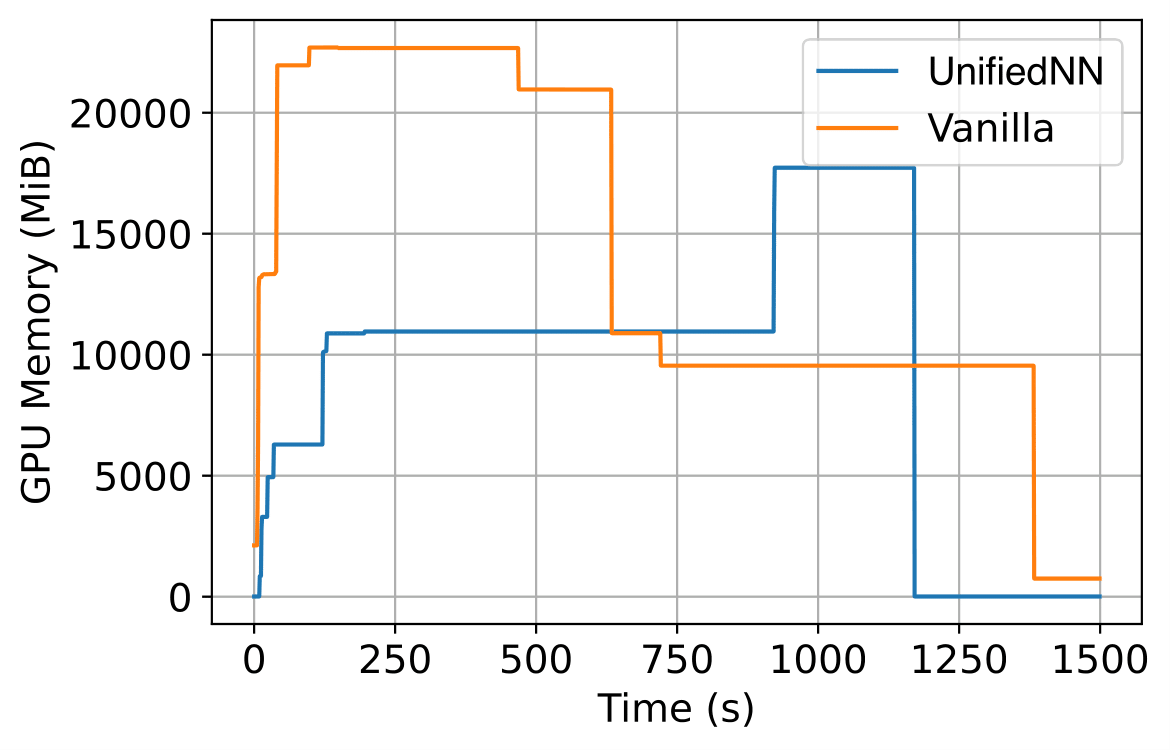}
         \caption{GPU memory usage}
     \end{subfigure}
        \caption{Comparison training time and GPU memory usage of the transformer model, XLM-RoBERTa, and the custom VGG16 when trained through vanilla PyTorch and \sol..}
        \label{fig:vgg_nlp_time}
\end{figure}


Figure~\ref{fig:vgg_nlp_time} presents the memory usage over time for vanilla PyTorch and \sol. In terms of GPU memory usage, our results demonstrate the same trend. Specifically, \sol used much less memory when training the models compared to vanilla PyTorch. When using vanilla PyTorch, the GPU memory usage peaked at 22697 MiB, whereas \sol peaked at 17726 MiB memory. As a result, \sol reduced memory consumption by 21\%.



\subsection{Evaluation of different scheduling algorithms}

\begin{figure}
     \centering
     \begin{subfigure}[b]{0.48\linewidth}
         \centering
         \includegraphics[width=\textwidth]{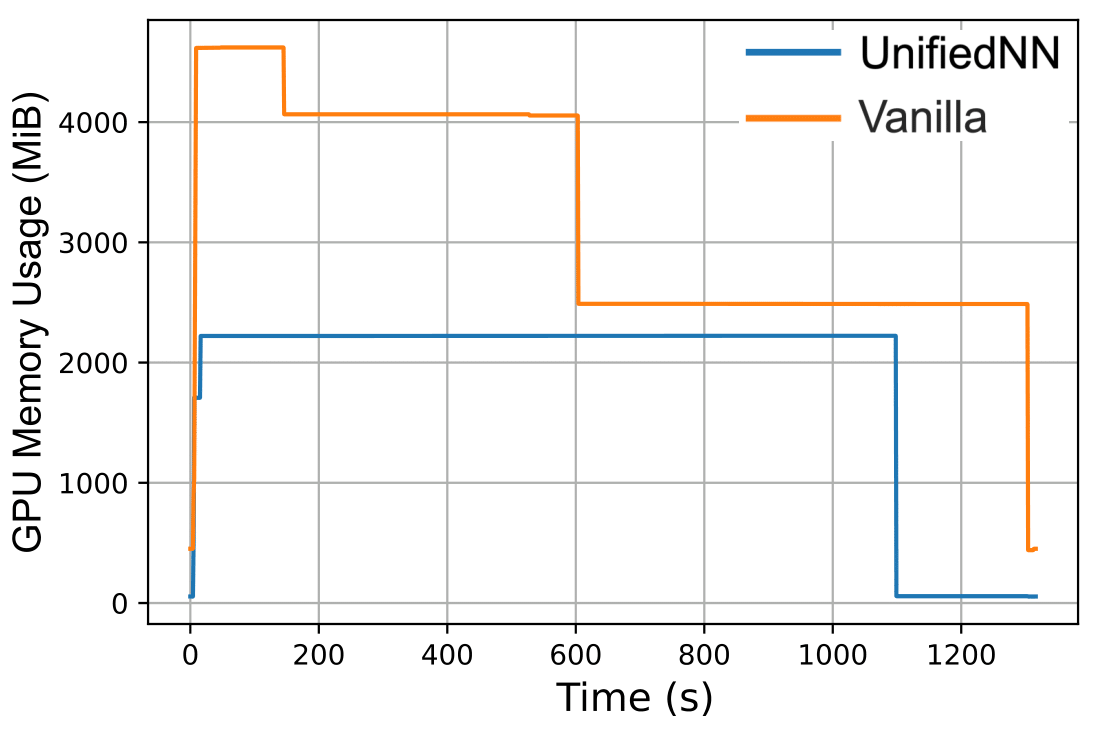}
         \caption{Training time}
     \end{subfigure}
          \begin{subfigure}[b]{0.48\linewidth}
         \centering
         \includegraphics[width=\textwidth]{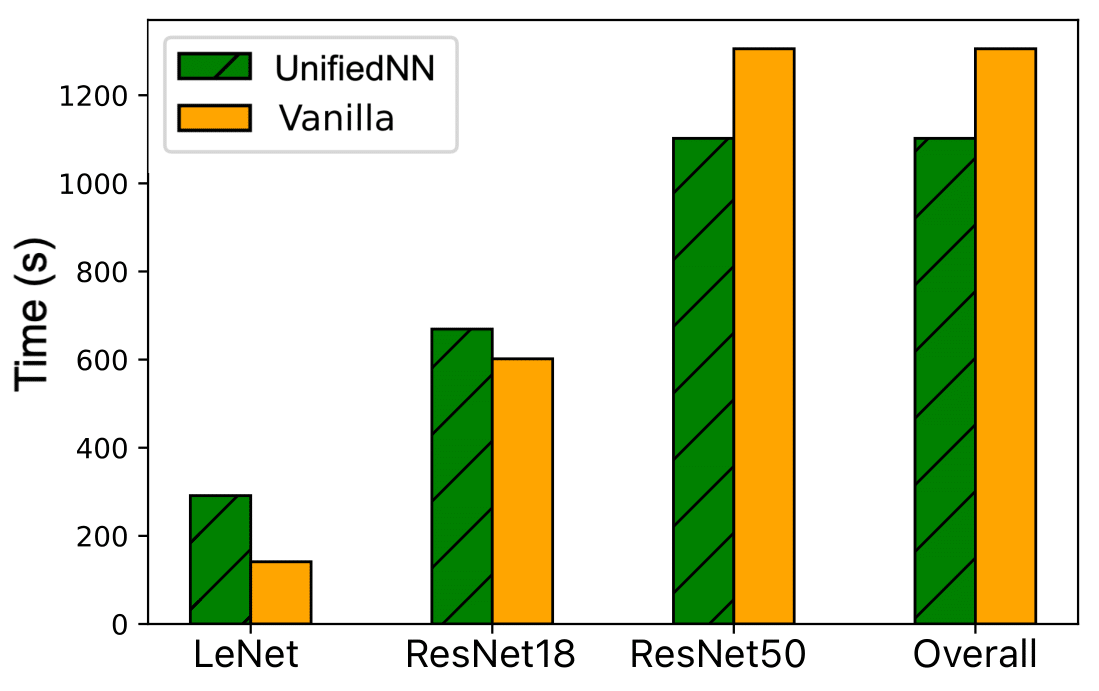}
         \caption{GPU memory usage}
     \end{subfigure}
        \caption{Comparison of vanilla PyTorch and \sol with the round robin algorithm.}
        \label{fig:rr_time}
\end{figure}

In this section, we evaluate the performance of different scheduling algorithms available in \sol. We also compare the performance of \sol to vanilla PyTorch to investigate appropriate use cases for each algorithm. As we have already established that \sol does not impact the accuracy of the models, we are only going to present results for the training time and the memory usage in this section. For this comparison, we trained the LeNet model with MNIST for 10 epochs, ResNet18 model with CIFAR10 for 25 epochs, and ResNet50 model with CIFAR100 for 50 epochs. Figures~\ref{fig:rr_time} present the memory usage and training time using the round-robin algorithm. Comparing the training times, we can see that with the Round Robin scheduling option, it took LeNet about 5 minutes to complete, which is 150 seconds more than the vanilla PyTorch taking about 2 and a half minutes. However, ResNet18 and ResNet50 took about 11 minutes and 18 minutes to complete, respectively, as compared to vanilla PyTorch, which took 10 minutes and 21 minutes to complete, respectively. Our results indicated that the smaller LeNet model suffered the most while the bigger ResNet50 models benefited by about 15\%. As a result, \sol performed 15\% better overall. The current implementation of \sol does not support preemption; as such, smaller models will suffer if combined with larger models that take more time to train per epoch. The memory allocation graph in Figure~\ref{fig:rr_time} suggests that the vanilla PyTorch peaked at 4621 MiB whereas \sol only peaked at 2223 MiB, which is a 51\% reduction in memory allocation. As our results indicate, the round robin scheduling algorithm is the right choice when merging and training NN models of similar sizes.

\begin{figure}
     \centering
     \begin{subfigure}[b]{0.48\linewidth}
         \centering
         \includegraphics[width=\textwidth]{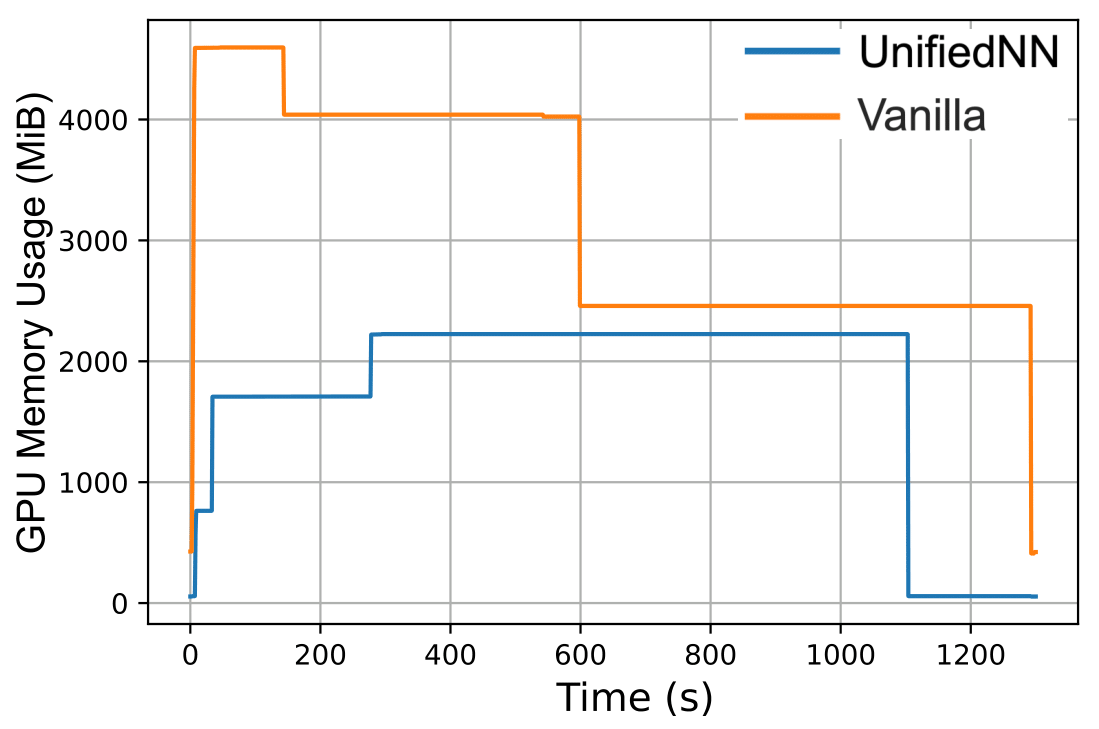}
         \caption{Training time}
     \end{subfigure}
          \begin{subfigure}[b]{0.48\linewidth}
         \centering
         \includegraphics[width=\textwidth]{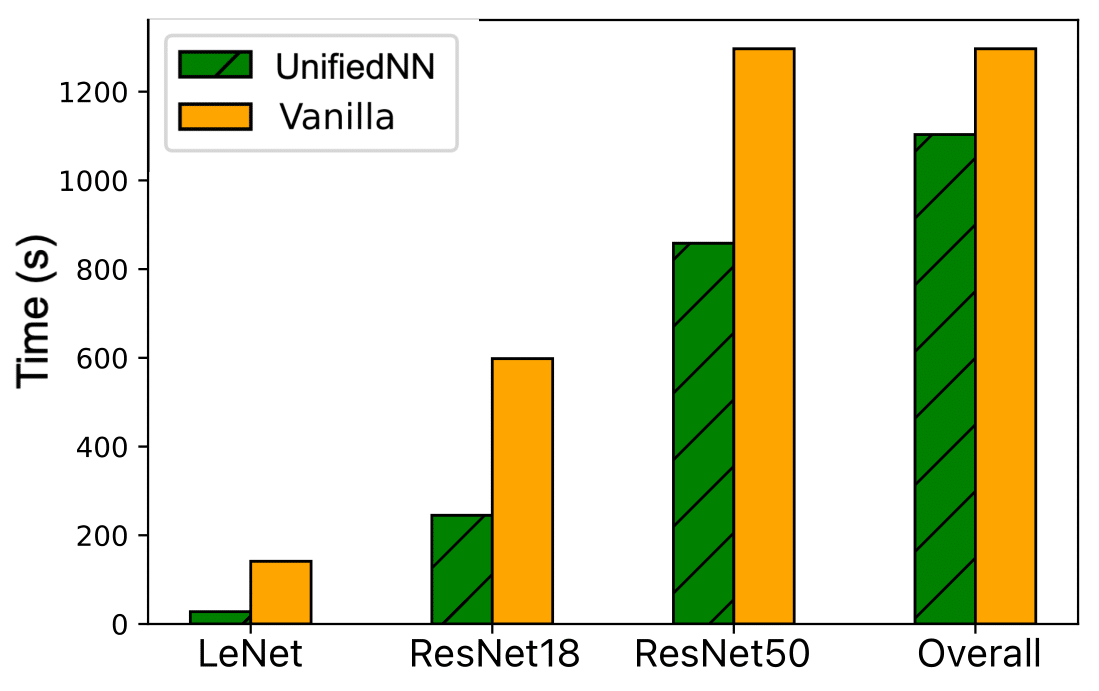}
         \caption{GPU memory usage}
     \end{subfigure}
        \caption{Comparison of vanilla PyTorch and \sol with the SJF algorithm.}
        \label{fig:sjf_time}
\end{figure}

Next, we evaluated the shortest job first algorithm. The definition of the length of a job can be either set using the model size or the number of epochs that the model needs to be trained. Here, we have used the number of epochs to determine the shortest sob. As a result, the LeNet model was trained first this time with a much lower training time. ResNet18 was trained next followed by ResNet50. We present the memory consumption and training time results in Figure~\ref{fig:sjf_time} respectively. Our results indicate that the training of LeNet took 27 seconds to complete, which is about 2 minutes faster than vanilla PyTorch taking about 2 and a half minutes. Moreover, ResNet18 and ResNet50 took 4 and a half minutes and 14 and a half minutes to complete, respectively, which are about 54\% and 33\% faster than vanilla PyTorch. Overall, \sol was about 15\% faster. The memory allocation raised slowly according to the requirements of the models and peaked at 2225 MiB, which is again 51\% lower than vanilla PyTorch. As such, the total memory allocation is much lower in this case as well.

\begin{figure}
     \centering
     \begin{subfigure}[b]{0.48\linewidth}
         \centering
         \includegraphics[width=\textwidth]{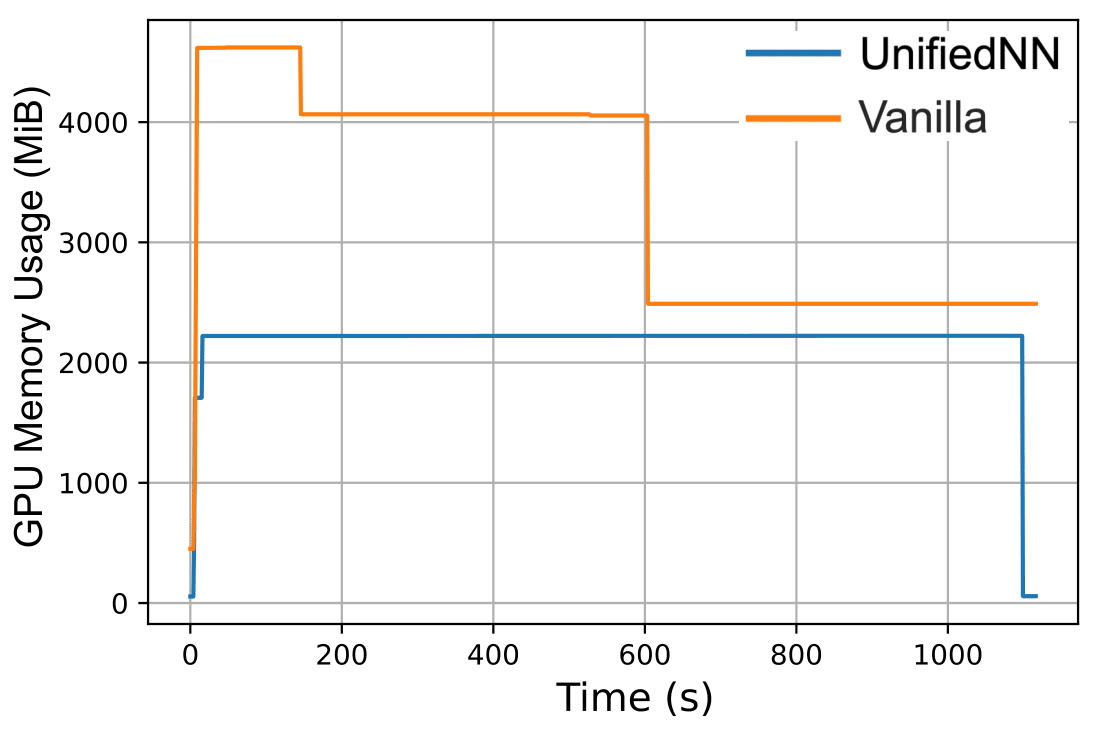}
         \caption{Training time}
     \end{subfigure}
          \begin{subfigure}[b]{0.48\linewidth}
         \centering
         \includegraphics[width=\textwidth]{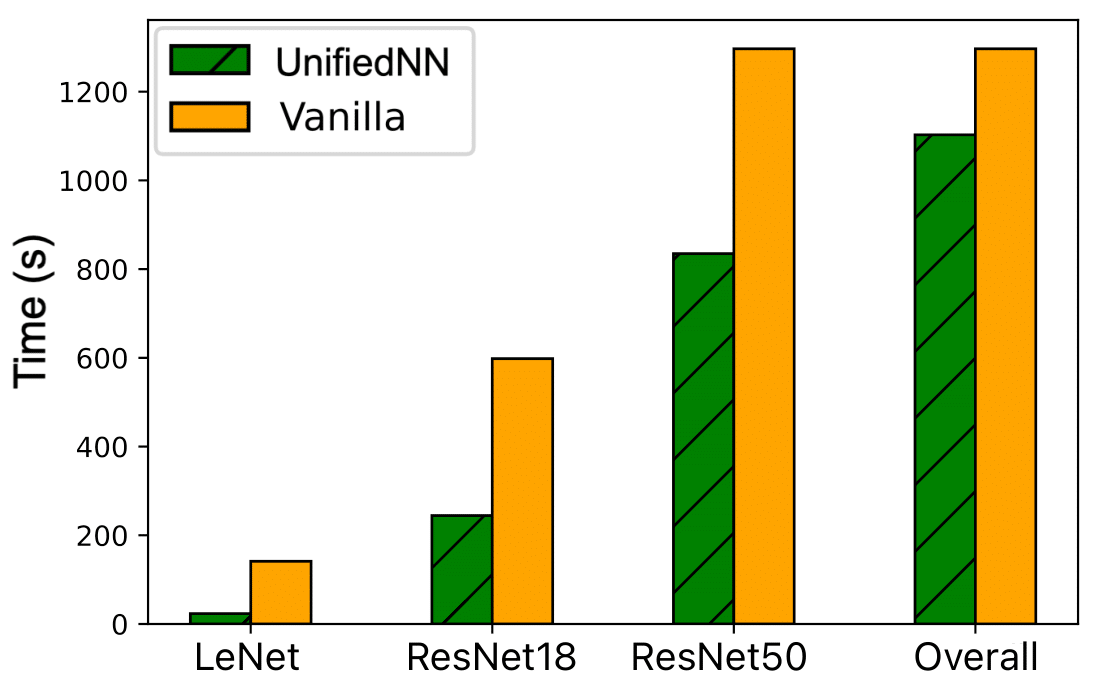}
         \caption{GPU memory usage}
     \end{subfigure}
        \caption{Comparison of vanilla PyTorch and \sol with the priority scheduling algorithm.}
        \label{fig:priority_time}
\end{figure}

Finally, we have evaluated the priority scheduling algorithm where ResNet50 was given the highest priority, followed by ResNet18 and LeNet, respectively. As such, ResNet18 finished first and LeNet finished last. This scheduling is useful for cases where there is a priority to train a certain (small or large) model as soon as possible. Figure~\ref{fig:priority_time} presents the training time results using the priority scheduling option. With this scheduling algorithm, ResNet50 only took about 14 minutes seconds, ResNet18 took 4 minutes, and LeNet took 14 seconds. Overall, it took about 18 and a half minutes, which is about 15\% faster than vanilla PyTorch. From Figure~\ref{fig:priority_time}, we can see that the memory utilization is almost the same as the round robin scheduling, which is 51\% lower than vanilla PyTorch. Specifically, we observe constant memory usage because the memory does not free up immediately.

Note that we also experimented with the first come first served algorithm and the results show the same trend as the results that we presented above. Overall, \sol will conserve GPU memory regardless of the selected scheduling algorithm. The different scheduling algorithm options allow service providers to prioritize user requests for training based on their own criteria. 


\subsection{Comparison with state-of-the-art frameworks}

\begin{figure}
     \centering
     \begin{subfigure}[b]{0.48\linewidth}
         \centering
         \includegraphics[width=\textwidth]{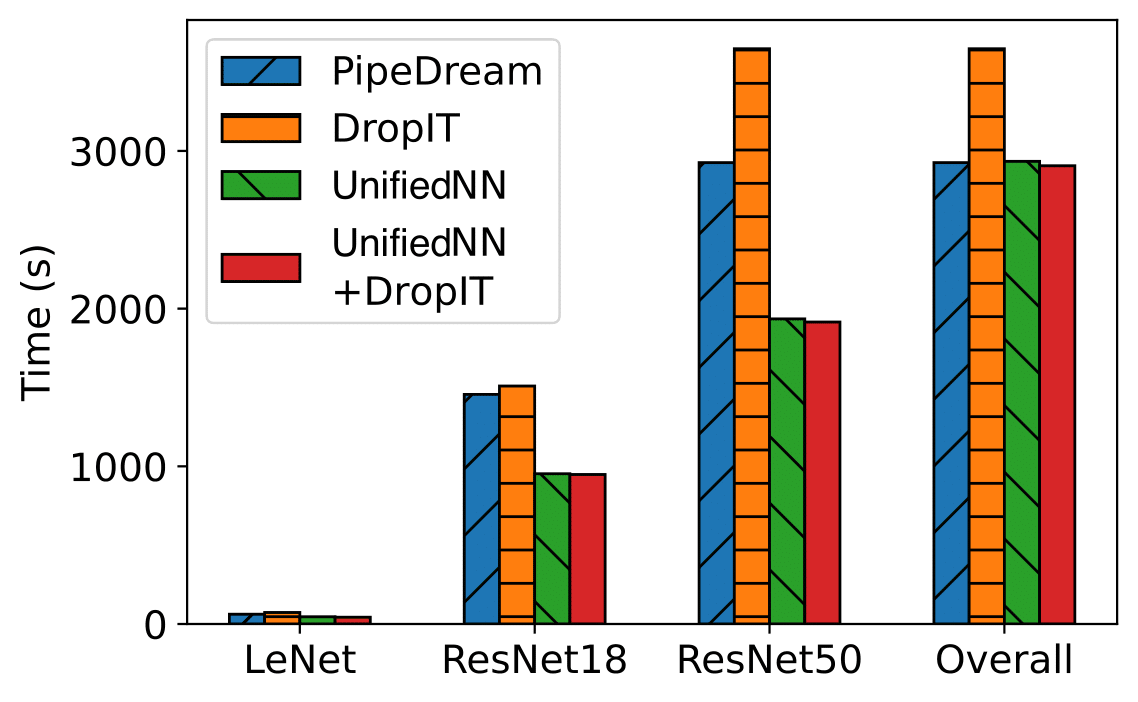}
         \caption{Training time}
     \end{subfigure}
          \begin{subfigure}[b]{0.48\linewidth}
         \centering
         \includegraphics[width=\textwidth]{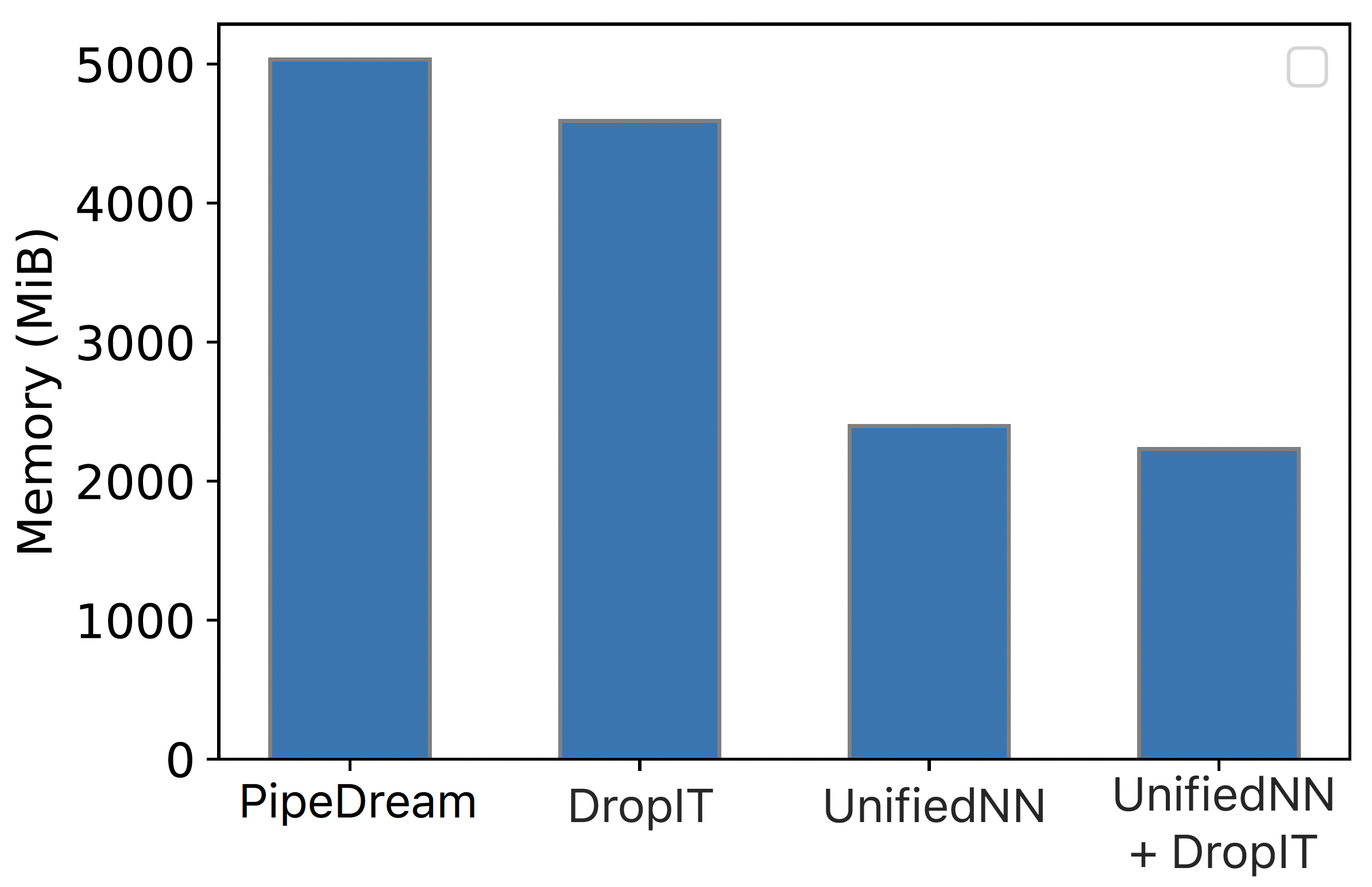}
         \caption{GPU memory usage}
     \end{subfigure}
        \caption{Comparison of\sol with PipeDream and DropIT.}
        \label{fig:frame_time}
\end{figure}

In this section, we have evaluated the performance of \sol in comparison with two state-of-the-art frameworks, PipeDream~\cite{pipedream} and DropIT~\cite{chen2022dropit}, which have been recently proposed. We trained LeNet, ResNet18, and ResNet50 concurrently with PipeDream, DropIT, and \sol using the hyper-parameters from Table~\ref{tab:eval_models}. We have also considered a case of implementing \sol with DropIT. Figure~\ref{fig:frame_time} presents the peak memory consumption of different frameworks while training the models mentioned above. Our results indicate that PipeDream peaked at 5034 MiB, DropIT peaked at 4592 MiB, and \sol peaked at 2396 MiB, which is 52\% lower than PipeDream and 48\% lower than DropIT. By implementing DropIT with \sol, we got a peak memory utilization of 2230 MiB, which is 7\% better than vanilla \sol. Since PipeDeam and DropIT are not designed to train multiple NN models, they did not perform well compared to training a single model.

Figure~\ref{fig:frame_time} also presents the training time of each model for the aforementioned frameworks. Our results indicate that the training of LeNet took 62 seconds with PipeDream, 73 seconds with DropIT, 46 seconds with \sol, and 43 seconds when we combine \sol with DropIT. \sol combined with DropIT was the fastest for LeNet, which is 30\%, 41\%, and 6\% better than PipeDream, DropIT, and vanilla \sol, respectively. For ResNet18, the training process took about 24 and a half minutes with PipeDream, about 42 minutes with DropIT, about 16 minutes with \sol, and about 15 and a half minutes with \sol combined with DropIT. \sol combined with DropIT performed 34\% faster than PipeDream and 37\% faster than DropIT for ResNet18. Finally, for ResNet50, the training process took about 49 minutes with PipeDream, about an hour with DropIT, 32 and a half minutes with \sol, and 32 minutes with \sol combined with DropIT, which are about 34\%, 47\%, and 1\% faster, respectively. Overall, \sol combined with DropIT was about 1\% faster than PipeDream, 20\% faster than DropIT, and 1\% faster than vanilla \sol while consuming 52\% less GPU memory.

\section{Discussion}
\label{sec:discussion}

\noindent \textbf{Limitations and extensions of the \sol design:} Our results demonstrate that \sol can significantly reduce the amount of used memory when training multiple models at a time. In addition, \sol reduces the training time itself in most cases. The performance of \sol can be further improved by adopting additional memory optimization strategies. Moreover, \sol currently supports four scheduling algorithms. Integrating more sophisticated scheduling algorithms with \sol can offer more flexibility in terms of scheduling training jobs under user/service constraints. \sol has also been designed under the assumption that multiple requests for NN model training will be available, so that \sol can combine and train these models at the same time. In other words, \sol has not been designed with the goal of training a single (individual) model. Finally, the current implementation of \sol is based on PyTorch. Nevertheless, the memory optimization mechanisms that \sol adopts are framework independent. As such, \sol can be implemented to work with TensorFlow and MXNet as well.


\noindent \textbf{Security and privacy considerations:} Although \sol combines multiple NN models, it does not leak information about a sub-model to other sub-models as long as the service provider is legitimate. In other words, the training process of a sub-model does not impact or interfere with the training process of other sub-models. In the case of multiple users with models trained in the context of the same hybrid model, a user does not have access to information about the models or datasets of other users. 



\section{Conclusion and Future Work}
\label{sec:conclusion}

In this paper, we presented \sol, a framework tailored towards efficient NN model training in a cloud environment. \sol reduces GPU memory usage and reduces the I/O overhead, thus reducing the overall time needed for model training. \sol performs particularly well when training multiple NN models simultaneously. \sol offers different scheduling algorithms to meet the needs of different cloud services and applications. We implemented and evaluated a prototype of \sol for various NN models that represent a wide range of use cases and applications. Our results demonstrate that \sol reduces memory consumption by up to 53\% and training time by up to 81\% when compared with vanilla PyTorch without impacting the model training and testing accuracy. In addition, our results indicate that \sol reduces memory consumption by up to 52\% and training time by up to 41\% when compared to state-of-the-art frameworks. 

In the future, we plan to explore the following directions: (1) implement and evaluate additional scheduling algorithms to accommodate the requirements of a broader spectrum of use cases; (2) explore additional mechanisms to optimize memory usage and reduce training time; and (3) work with cloud providers to deploye and evaluate \sol in the context of an actual cloud computing infrastructure.


\bibliography{aaai25}

\end{document}